\documentclass[twocolumn]{svjour3}         
\pdfoutput=1 

\smartqed  
\usepackage{graphicx}
\usepackage{multirow}
\usepackage{amsmath}
\usepackage{color}

\def\ie{\emph{i.e.,\,\,}} 
\def\eg{\emph{e.g.,\,\,}} 
\def\etal{\emph{et. al.\,\,}}

\begin{document}

\title {Incorporating Near-Infrared Information~ \newline into Semantic Image Segmentation
}
\subtitle{}


\author{Neda~Salamati$^{1}$,
        Diane~Larlus$^2$,
        Gabriela~Csurka$^2$,
        and~Sabine~S\"{u}sstrunk$^1$
}
\institute{$^1$ IVRG, IC, \'{E}cole Polytechnique F\'{e}d\'{e}rale de Lausanne, Switzerland\\
$^2$ Xerox Research Centre Europe (XRCE), Meylan, France\\
}
\date{Received: date / Accepted: date}
\authorrunning{Neda~Salamati,
  Diane~Larlus,
  Gabriela~Csurka,
  and~Sabine~S\"{u}sstrunk}

\maketitle

\begin{abstract}
Recent progress in computational photography has shown that we can acquire
near-infrared (NIR) information in addition to the normal visible (RGB) band,
with only slight modifications to standard digital cameras. Due to the proximity
of the NIR band to visible radiation, NIR images share many properties with
visible images. However, as a result of the material dependent reflection in the
NIR part of the spectrum, such images reveal different characteristics of the
scene. We investigate how to effectively exploit these differences to improve
performance on the semantic image segmentation task. Based on a state-of-the-art
segmentation framework and a novel manually segmented image database (both
indoor and outdoor scenes) that contain 4-channel images (RGB+NIR), we study how
to best incorporate the specific characteristics of the NIR response. We show
that adding NIR leads to improved performance for classes that correspond to a
specific type of material in both outdoor and indoor scenes. We also discuss the
results with respect to the physical properties of the NIR response.

\end{abstract}

\section{Introduction}

\begin{figure}
\begin{tabular}{ccc}
\hspace{-0.3cm}\includegraphics[height=1.8cm]{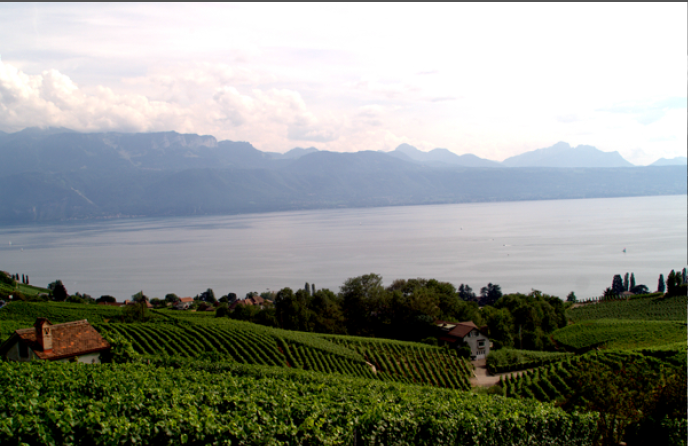}&
\hspace{-0.3cm}\includegraphics[height=1.8cm]{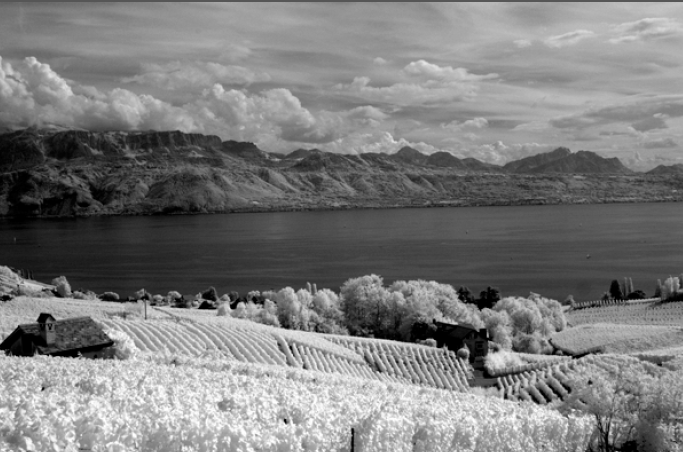}&
\hspace{-0.3cm}\includegraphics[height=1.8cm]{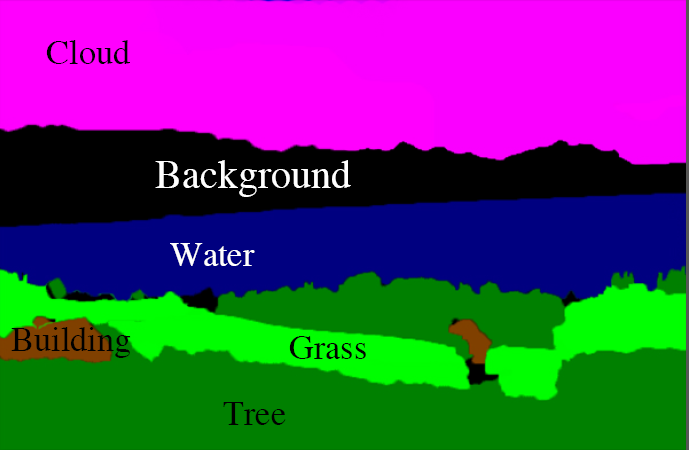}\\
\hspace{-0.3cm} \tiny RGB & \hspace{-0.3cm} \tiny NIR & \hspace{-0.6cm} \tiny 
Groundtruth segmentation
\end{tabular}
\caption{The semantic
  segmentation task consists in labeling every pixel of an image as belonging to
  one of the classes of interest (\eg: $grass$, $sky$, $tree$, $water$)  or to the
  background. Here, we demonstrate that semantic segmentation can be improved
  with additional near-infrared information.}
\label{fig:semseg_principle}
\end{figure}

In computer vision, semantic image segmentation is the task that assigns a
semantic label to every pixel in an image. For example, an outdoor image could
be segmented into different regions (\ie groups of pixels) that correspond to
the labels $grass$, $sky$, $tree$, and $water$ (see
Figure~\ref{fig:semseg_principle} for an example). Semantic segmentation thus
necessitates the joint recognition and localization of several classes of
interest, that can be both object classes or background classes.

This task has received a lot of attention over the last decade, and many methods
have been proposed to semantically segment RGB
images~\cite{shotton2006,verbeek2007,schmid2008,larlus2010,csurka2011,koltun2011}. Most
proposed models combine two components: (i) a local recognition process that identifies
the classes and (ii) a regularization process that groups the pixels into
semantic regions, often using contour information. Promising results have been obtained
over a broad range of categories and for a large variety of scenes, but the
problem is far from being solved.
Segmentation algorithms tend to fail in the presence of a cluttered background
or distractor objects, as well as when the objects to be segmented are composed
of several colors or patterns. Figure~\ref{fig:failing_samples}-a shows
instances of the $cup$ class that are difficult to segment. Because of surface
color variations, even for classes with a consistent color, such as
$vegetation$, relying solely on RGB-based features can be insufficient as the
color can drastically vary under different lighting conditions.

All these difficulties challenge both components of the segmentation models.
The recognition part that learns an appearance model of categories based solely
on RGB values has issues with objects where both the color and the texture
significantly vary within the object classes, which is the case for most
man-made objects.  Concerning the regularization part, object contours can be
easily confused with strong edges originating from texture patterns on the
object, background clutters or any distracting regions.

One way to overcome these issues is to incorporate additional information within
the segmentation process. It was shown for example in \cite{Crabb2008} that 
adding depth information when available (\eg obtained with 
stereo cameras or using range sensor data) can significantly improve segmentation
results. Other methods have considered multi-spectral data to enhance
segmentation~\cite{Blaschke2010}. However, these methods require
costly and specific acquisition equipment.

In this paper, we consider additional information that can be obtained from a
standard digital camera: the near-infrared part of the spectrum that is captured
by consumer camera sensors. We study this near-infrared channel in addition to
RGB images in the context of semantic image segmentation. This paper extends the
work presented in~\cite{Salamati2012} by studying a wider range of scene types,
and a deeper analysis of the results. We use the same state-of-the-art
segmentation framework as~\cite{Salamati2012}. To evaluate the proposed
approach, first we contribute with a dataset that extends the ones
of~\cite{brown2011,Salamati2012} with pixel-level annotation of 770
registered RGB and NIR image pairs\footnote{This annotated dataset is available
  at http://ivrg.epfl.ch/research/infrared/dataset} for 23 semantic classes. In order to
discuss the advantages that NIR brings to each class of material, the dataset is
split into indoor and outdoor scenes and the performance is reported on each
subset separately. The outdoor dataset contains 10 different classes, and the
indoor dataset contains 13 classes.

Our second contribution is an in-depth study of the results obtained by our
proposed model. We extended a state-of-the-art segmentation framework with
different strategies for incorporating the NIR channel that improve over
conventional RGB-only segmentation results. This framework is based on a
conditional random field (CRF), where we exploit different ways to combine the
visible and NIR information in the recognition part and in the regularization
part of the model. Based on the segmentation results, the paper fully discusses
the accuracy obtained for each class of material, together with success and
failure cases, and puts them in perspective with the material characteristics of
the NIR radiation. In particular, we show that the overall improvement is due to
a large improvement for certain classes whose response in the NIR domain is
particularly discriminant.

The rest of this article is organized as follows. Section~\ref{sec:nirimaging}
reviews the relevant literature on NIR imaging while
Section~\ref{sec:semanticsegmentation} discusses different approaches to
semantic segmentation. Section~\ref{sec:model} describes our semantic image
segmentation system.  The experimental setup and evaluation procedure are
explained in Section~\ref{sec:experiment} and experimental results are exposed
and discussed in Section~\ref{sec:results}
and~\ref{sec:discussion}. Section~\ref{sec:conclusion} concludes the paper.

\section{Near-infrared (NIR)}
\label{sec:nirimaging}

\subsection{NIR properties and their relevance to segmentation}

\begin{figure}[t]
  \centering
  \includegraphics[width=7.5cm]{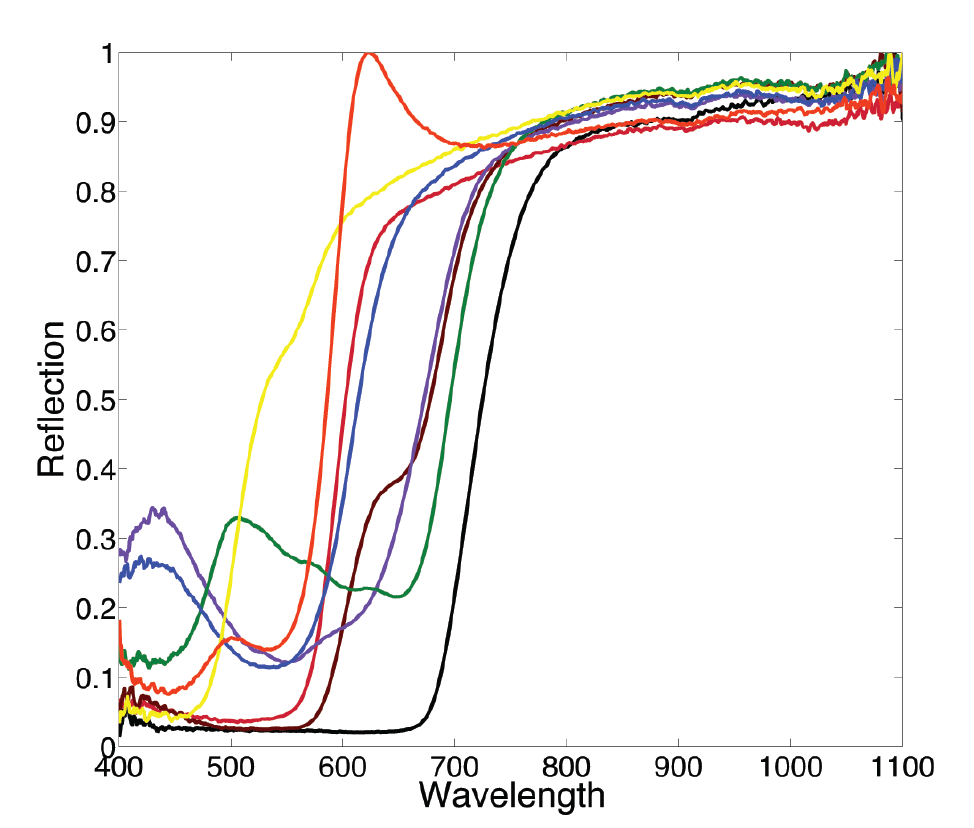}
  \caption{Spectral reflectance of 20 different fabrics. The reflection of these
    samples is different in the visible part of the spectrum, which for a given
    camera and lighting condition leads to different color values. However, as
    these samples belong to the same material, their spectral reflectance in the
    NIR range is not significantly different.}
  \label{fig:facspec}
\end{figure}

NIR spectra are influenced by the chemical and physical structure of different
material classes, which makes NIR suitable for material
classification~\cite{burns2001}. Figure \ref{fig:facspec} shows that for a given
material, regardless of the object color in the visible part (400-700~nm), the
reflection in the NIR band (700-1100~nm) remains the same.

\begin{figure}[t]
  \centering
  \includegraphics[width=\linewidth]{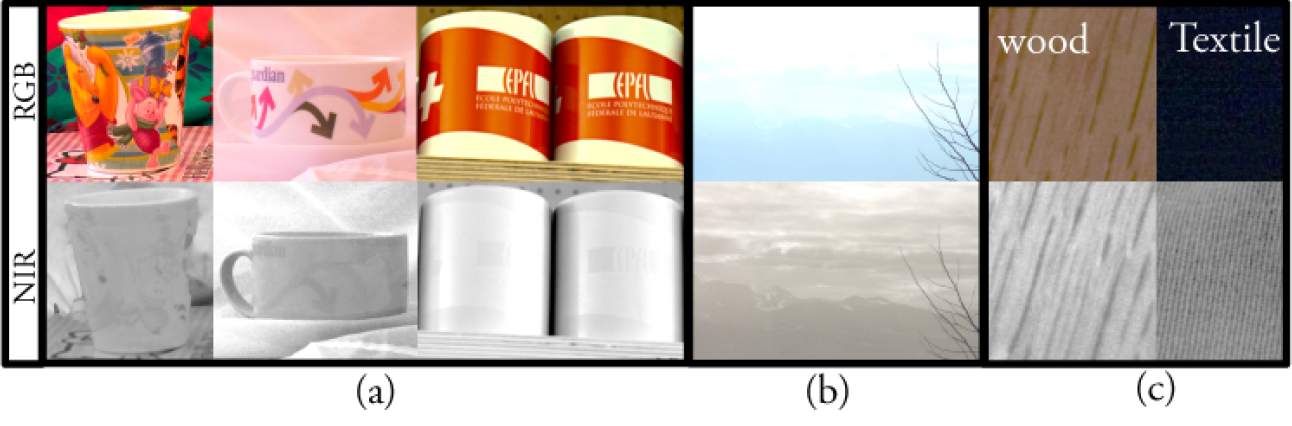}
  \caption{Challenging cases for RGB-only semantic segmentation. In NIR images (a) cups in different colors have the same brightness, (b) haze is transparent, and (c) texture is more intrinsic to the material.}
  \label{fig:failing_samples}
\end{figure}

\cite{salamati2009} shows that each class of material has an intrinsic behavior
in the near-infrared (NIR) part of the spectrum. The material dependency of NIR
reflection has been proven to be useful in low-level segmentation~\cite{salamati2010} 
and produces segments that correspond to changes of material~\cite{salamati2010}. 
Our previous works show that NIR could enhance scene classification~\cite{salamati2011}
and semantic segmentation as well~\cite{Salamati2012}.

The properties of infrared images~\cite{Morris2007} have been used by the remote
sensing~\cite{zhou2009,walter2004} and military~\cite{dowdall2005} communities
for many years to detect and classify natural and/or man-made objects. However,
in this paper we approach semantic image segmentation from a different point of
view. Unlike most remote sensing applications that use true hyper-spectral
capture with several bands in the NIR part and the IR part of the spectrum,
our framework only uses one  single channel that integrates all NIR
radiation. The single NIR channel can be captured by the standard sensor of any
digital camera. Moreover, in remote sensing and military applications, the focus is mostly
on aerial photography, forgery, and human detection, while  this work 
tackles semantic segmentation in everyday photography.

As already mentioned earlier, state-of-the-art semantic segmentation techniques
still encounter some difficulties in both recognizing different classes and
detecting the actual boundary of semantic objects. 
Such shortcomings are mostly due to four reasons : appearance variation,
cluttered backgrounds, illumination differences, and hazy atmosphere in outdoor
scenes.
Learning a color-based appearance model of classes based only on RGB values is very
challenging due to the appearance variation
of a single object, or the varying lighting conditions.
As another situation, the segmentation of $cloud$ and $mountain$ classes often
fail in a hazy atmosphere (see Figure~\ref{fig:failing_samples}-b for illustration).  
Texture has shown to be a powerful cue for semantic
segmentation~\cite{csurka2011,nils2009}. However, texture can be confused with
color patterns and other immediate surface impurities, as surface reflectance in
the visible part of the spectrum that is not intrinsic to the corresponding
object material (see Figure~\ref{fig:failing_samples}-c for illustration).

Light in the near-infrared range has physical properties intrinsically different
than in the visible range. The intensity values in NIR images are more
consistent across a single material and consequently across a given class
region. For instance, vegetation is consistently very bright, and sky and water
are very absorbent in the NIR band. Furthermore, a large number of colorants and
dyes are transparent to NIR. Thus, material-intrinsic texture properties are
easier to capture in this part of the spectrum.

Because of all these reasons, NIR is a great complement to the visible
information and improves the labeling of the semantic classes that correspond to
specific materials or textures. For instance, the class $cup$ is often made of
very specific classes of material (porcelain, ceramic or plastic) and regardless
of the pattern and color of the object, NIR has a unique response for each of
these materials (see Figure~\ref{fig:failing_samples}-a for illustration).
Another interesting aspect is the fact that atmospheric haze is transparent to
NIR, hence the borders of objects in hazy weather conditions is better defined in NIR
images.

\subsection{NIR imaging}
NIR imaging is used in different areas of science. 

NIR spectroscopy (NIRS) is employed for material identification and forgery
detection. NIRS is a non-destructive technique used to study the
interactions between incident light and material surfaces; it is based on the
assumption that surface reflection in the NIR band proves to be critical for detection
of different classes of material~\cite{burns2001,whelan2003}. The need for
little or no preparation of samples has made it one of the most used techniques
for material identification in industry~\cite{burns2001}.

In remote sensing, multi-spectral images are captured in order to detect,
characterize and monitor different regions, such as vegetation and soil. In
such applications, region reflection in both the visible and IR parts of the
spectrum is required~\cite{zhou2009,walter2004}. It has been shown that both
offer valuable information providing bio-signatures for different classes of
vegetation and soil properties~\cite{blackburn2007}.

In both remote sensing and NIR spectroscopy, hyper-spectral data is needed to
accurately identify material classes. However, these capturing devices are
highly technical and expensive, hence they are of limited usefulness outside
laboratory conditions.

\begin{figure}[t]
  \centering
  \includegraphics[width=8cm]{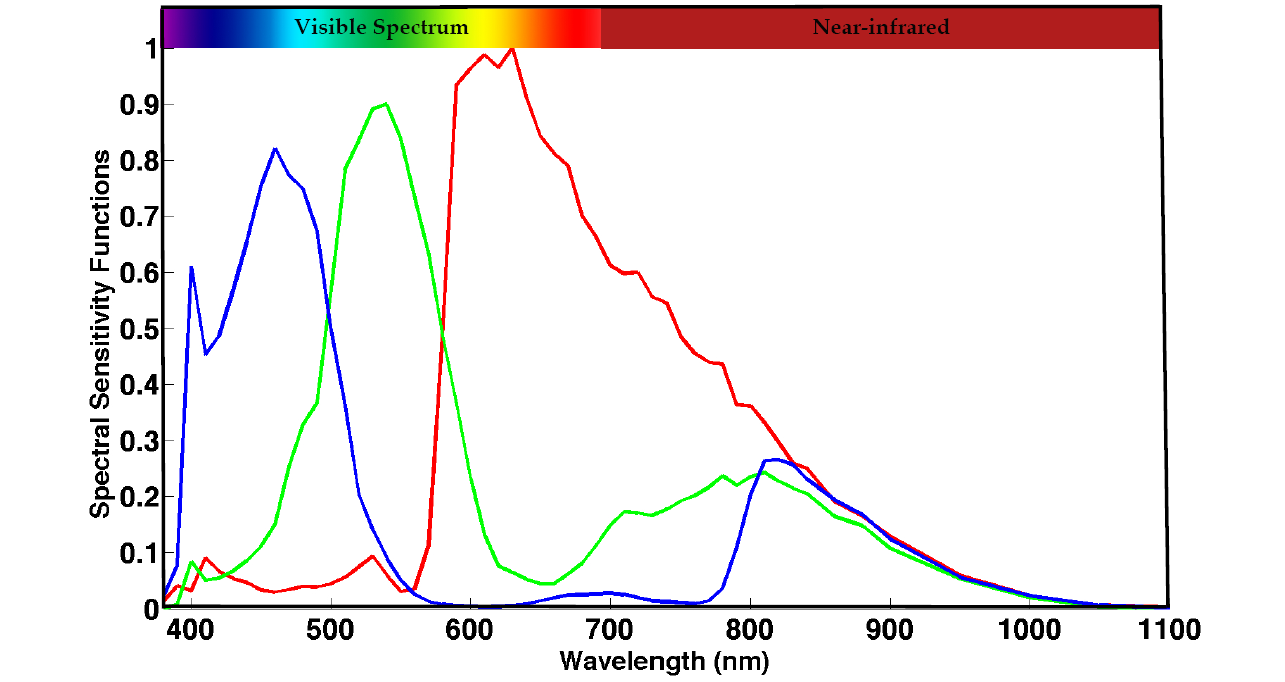}
  \caption{Typical transmittance curves of the RGB filters of the silicon sensor
    found in the NikonD90 camera.}
  \label{fig:ccd_sens}
\end{figure}

The most common sensors in consumer digital cameras, \ie CMOS, is made of
silicon and thus is intrinsically sensitive to wavelengths from roughly 350 nm
to 1100 nm. The material employed
to create the color filter array (CFA) is also transparent
to the NIR. Figure~\ref{fig:ccd_sens} shows the transmittance of the CFA
filters of a NikonD90 with the hot mirror (NIR-blocking filter that is normally
put by manufacturer) removed. Hence, if the
NIR-blocking filter is omitted from the camera, the sensor has the capability 
to capture both NIR and visible bands; no modification of the sensor is
required as shown in~\cite{fredembach2008}.

Currently, two main methods are used to jointly capture the NIR and visible
images. The first approach uses two cameras with a beam
splitter~\cite{zhang2008}. In the second approach, a camera with NIR and visible
filters is used to sequentially capture two images that are then
registered~\cite{fredembach2008}.  As a third method, Kermani~\etal designed a
new color filter array that could be used for demosaicing the ``full-spectrum"
raw data into a visible and NIR pair~\cite{kermani2011}. Although all the images
we used in our experiments have been acquired with the second approach, any of
these acquisition methods can be used with the proposed approach.

Several recent works have used such 4-dimensional images -- RGB and NIR channels
-- for standard computer graphics and computer vision tasks. In image
enhancement, we can mention haze removal~\cite{Schaul2009}, skin
enhancement~\cite{Fredembach2010}, dark-flash photography~\cite{Krishnan2009},
videoconferencing~\cite{hppeople}, and real-time 3-dimensional depth
imaging~\cite{kinektpeople}. In computer vision, the intrinsic properties of
material classes in NIR images make this information a relevant choice in image
scene classification~\cite{brown2011,salamati2011} or material-based
segmentation~\cite{salamati2010}.

\section{Semantic Segmentation}
\label{sec:semanticsegmentation}
Semantic image segmentation is the computer vision task that consists in
partitioning an image into different semantic regions, where each region
corresponds to a class. It can be seen as a
pixel-level categorization task.

The appearance of the classes of interest are learnt during a training step from
images annotated at the pixel-level. Learnt models are based on appearance
descriptors that are often based on Gaussian derivative filter outputs or
texture descriptors such as SIFT~\cite{lowe2004}, or based on colors
statistics~\cite{vandeweijer2006,clinchant07}. Appearance descriptors are
extracted at the pixel-level~\cite{shotton2006,shotton2008} or statistics are
extracted at the patch level, on a (multi-scale) grid of
patches~\cite{csurka2011,verbeek2007,larlus2010}, or at detected interest point
locations~\cite{yang07}. The location of the pixels is sometimes used as a cue
as well~\cite{shotton2006,shotton2008}. In general, low-level features are used
to build higher-level representations, such as texton
forests~\cite{shotton2008}, bags-of-visual-words
(BoVW)~\cite{larlus2010,verbeek2007}, or Fisher vectors~\cite{csurka2011,Hu2012}
that are fed into a classifier to predict class labels at pixel
level~\cite{shotton2008}, patch level~\cite{larlus2010,csurka2011} or region
level~\cite{yang07}.

The local appearance models are often combined with a local label consistency
component that adds constraints on neighboring
pixels~\cite{shotton2006,larlus2010} or within an image region
(super-pixel)~\cite{csurka2011,Hu2012}. Sometimes, global consistency also takes
into account the whole image~\cite{shotton2006,larlus2010} and/or the context of
the object class~\cite{ladicky2010}. State-of-the-art semantic segmentation
methods typically integrate these components into a unified probabilistic
framework such as a conditional random field (CRF)
\cite{verbeek2007,kohli2009,ladicky2009,kohli2010,larlus2010,Hu2012}, or these
components are learnt independently and used at different stages of a sequential
pipeline~\cite{yang07,csurka2011}.

In this paper, we consider the former and use a CRF model that combines local
appearance predictions with local consistency of labels. Label consistency is
enforced by a set of pairwise constraints between neighboring pixels. We used a
standard contrast-sensitive Potts model. Note that more complex CRF models were
proposed that incorporate object co-occurrences \cite{ladicky2010} or object
detectors~\cite{ladicky2010and}, use fully connected CRF
models~\cite{koltun2011}, hierarchical associative CRFs \cite{ladicky2009} or
higher order potentials~\cite{kohli2009,kohli2010}.  The aim of our paper is to
evaluate the potential advantage of the NIR information in a standard semantic
image segmentation model, so we used a very standard CRF model and we do not
consider these more complex models. We believe that the conclusions of this
paper would extend to these models and that this constitutes potential
extensions of the presented framework. Comparing these methods is beyond the
score of this paper.

\section{Our CRF framework}
\label{sec:model}

We represent the label of a pixel
 $i$ with a random variable $X_i$ 
taking a value from the set of labels $\mathcal{L}= \{l_1, \dots, l_n\}$, $n$
being the number of classes. Let $X$ be the set of random variables representing 
the labeling of an image and 
$X=\textbf{x}$ an actual ($i.e.$, possible) labeling.  The posterior distribution 
$P(X=\textbf{x}\mid \textbf{D})$, given the observation \textbf{D}
over all possible labeling of a CRF,  is a $Gibbs$ distribution 
and can be written as:
\begin{equation}
P(X=\textbf{x}\mid \textbf{D}) =\frac{1}{Z}\exp (\; - \underbrace{\sum_{c\in C}
  \psi_c(\textbf{x}_c)}_{E(\textbf{x})} \; )
\end{equation}
where $\psi_c(\textbf{x}_c)$ are potential functions over the variables
$\textbf{x}_c = \{ x_i,i\in c\}$ and $Z$ is a normalization factor. In this
equation, a clique $c$ defines a set of random variables $X_c\subseteq X$ that 
depend on each other, and $C$ is the set of all cliques. 
Please note that the observation \textbf{D} represents the actual pixel values in the image.
Accordingly, the $Gibbs$ energy is defined as
\begin{equation}
E(\textbf{x}) = -\log (P(X=\textbf{x}\mid \textbf{D})) - \log Z 
\end{equation}
Using these notations, the goal of semantic segmentation is to find the most probable labeling \textbf{x}$^*$, which is defined as the maximum a posteriori (MAP) labeling:
\begin{eqnarray*}
\textbf{x}^*  =   \textrm{argmax}_{\textbf{x} \in \mathcal{L}^N} \,
P(X=\textbf{x}\mid \textbf{D}) =
 \textrm{argmin} _{\textbf{x} \in \mathcal{L}^N}\,E(\textbf{x})
 \label{eq:energy}
\end{eqnarray*}
where $N$ is the number of pixels.

In our  CRF model, the energy function $E$ is composed of two terms,
a unary potential $E_{un}$  and a pairwise potential $E_{pair}$ \footnote{As discussed, more complex models can contain higher
  order potentials}. 
The unary term is responsible for the recognition part of the model and the
pairwise term encourages neighboring pixels to share the same label. 
We assign a weight $\lambda$ to
$E_{pair}$ that models the trade-off between recognition
and spatial regularization. More formally,
\begin{align}
E(\textbf{x}) &= E_{un}(\textbf{x}) + \lambda E_{pair}(\textbf{x})
\nonumber \\
& = \sum_{i \in \nu} \psi_i(x_i) + \lambda \sum_{(i,j) \in \varepsilon} \psi_{i,j}(x_i,x_j) 
\label{eq.fullcrf}
\end{align}
where $\nu$ corresponds to the set of all image pixels and $\varepsilon$ the set
of all edges connecting the pixels $i,j \in \nu$. Usually, 4-neighborhood or
8-neighborhood systems are considered. We used the former.

Usually, these models consider unary and pairwise potentials that are built
using information extracted from RGB images. In the following, we show
how to extend both the unary and the pairwise potential by integrating the NIR
channel in the above energy term.

\subsection{The unary term} 
\label{sec:unary}

The unary part of the CRF ($E_{un}$) is defined as the 
negative log likelihood of a label being assigned to pixel $i$. 
It can be computed from the local appearance model for each class. 
\begin{eqnarray*}
E_{un}(\textbf{x}) = \sum_{i \in \nu} \psi_i(x_i) =  \sum_{i \in \nu} -\log (P(X_i=\textbf{x}_i \mid \textbf{D}))
\end{eqnarray*}

Although the CRF uses pixels, the recognition could be made at several levels. We
could describe pixels directly, or work with patches extracted following a
regular grid. To build our local appearance model
we follow \cite{csurka2011} and use  patch-level 
Fisher Vector (FV) representations.
FVs~\cite{Perronnin2010} encode higher order statistics than the visual word counts in the
Bag-of-Visual-Words (BoVW) representation~\cite{csurka2004}. We chose to use FV as they outperform the
BoVW (as shown in \cite{csurka2011}), and as they are highly competitive for
object classification even with linear classifiers~\cite{chatfield2011}.
However, any similar representation could also have been used
for the study.

In a nutshell, our approach is the following: We extract overlapping 
image patches on a multi-scale grid and describe them with  
low-level descriptors. The dimension of  
these features is reduced using 
principal component analysis (PCA) before building a  Gaussian mixture 
model (GMM) based  visual codebook that allows to 
transform the low-level representation of each patch into a 
FV (see \cite{Perronnin2010,csurka2011} for more details). 
For each class, we train a patch-level linear classifier using strongly labeled
training images (\ie segmented images), and the classification score of each
patch in a test image is transformed into a probability. The class posterior
probabilities at the pixel level are obtained as a weighted average of the patch
posteriors, where the weights are given by the distance of the pixel to the
center of the patch as in~\cite{csurka2011}.

Several types of features can be used to describe patches.  The most popular
descriptors for RGB (=visible) images are color and texture features. Here, we
consider the popular SIFT~\cite{lowe2004} feature to describe the local texture
and local color statistics~\cite{clinchant07} to describe the color.  The
latter, referred to as $COL$, encodes the mean and standard deviation of the
intensity values in each image channel for each cell of a 4x4 grid covering the
patch (same cells as in the case of $SIFT$). In our visible-baseline approach,
these color statistics are computed on the R,G and B channels, hence we will
denote their concatenation by $COL_{rgb}$.

SIFT encodes local texture with a set of histograms of oriented gradients
computed on a 4x4 grid covering the patch. In general, it is computed on the
patch extracted from the luma channel of the visible RGB image that can be
approximated by $L = 0,299 R + 0,587 G + 0,114 B$.  It will therefore be denoted
by $SIFT_l$.

SIFT is sometimes extended by computing histograms of oriented gradients in
each color channel and by concatenating the obtained histograms. This
descriptor is called multi-spectral SIFT and denoted by $SIFT_{rgb}$.

When an additional NIR channel is considered, we can also extract the corresponding
color and SIFT descriptors. We will denote the additional ones by $COL_{n}$
and $SIFT_n$, respectively. We can concatenate them with features extracted from
the standard RGB image leading to \eg $COL_{rgbn}$ or $SIFT_{rgbn}$.

Due to the high correlation of RGB and NIR channels, \cite{brown2011} shows that
incorporating NIR information in a de-correlated space improves the performance
of image classification. We make use of the same idea, decorrelating the
4-dimensional RGB-NIR color vector by performing Principal Component Analysis
(PCA). In this alternative PCA space, we consider $COL_{p1234}$ and
$SIFT_{p1234}$ (see~\cite{brown2011,salamati2011} for more details).

In Section~\ref{sec:results}, we compare and discuss the performance of using each of these 
descriptors in the unary term of our energy function.

\subsection{Pairwise term} 
\label{sec:binary}

The pairwise term  $E_{pair}$ of our CRF takes the form of a contrast sensitive Potts model:

\begin{align}
E_{pair}(\textbf{x}) &= \sum_{(i,j) \in \varepsilon} \psi_{i,j}(x_i,x_j)
\nonumber \\
&= \sum_{(i,j)\in \varepsilon} \bar\delta_{x_i,x_j}\exp (-\beta \parallel q_i-q_j\parallel
^2) 
\end{align}
where $\bar\delta_{x_i,x_j}=1$ if $x_i \neq x_j$ and $\bar\delta _{x_i,x_j}=0$
otherwise. We set $\beta = ({2< \parallel q_i- q_j \parallel ^2 >})^{(-1)}$, as
in the work of~\cite{rother2004}. This
potential penalizes disagreeing labels in neighboring pixels, and the penalty is
lower where the image intensity changes. In this way, borders between 
predicted regions are encouraged to follow image edges.

In general, the pixel values $q_i$ in the Potts model
correspond to the RGB values of pixels.
When this is the case we denote the pairwise term by 
$VIS$, as it corresponds to the visible baseline that uses only the RGB image.
However, when NIR information is available, we can use the NIR channel in the
pairwise potential. In this case, $q_i$ corresponds to the intensity of the
pixel in the NIR channel, and the potential will be denoted by $NIR$. Finally,
when the Potts model uses the intensity values from the 4 channels ($q_i$ is 4
dimensional), the pairwise potential is denoted by $VIS + NIR$.

\subsection{Model inference} 
Given the CRF model defined in Eq~(\ref{eq:energy}), we want to find the most
probable labeling ($\textbf{x}$*), \ie the labeling that maximizes the posterior
distribution or that minimizes our energy function. This is a NP-hard problem
for many practical multi-label computer vision problems and approximation
algorithms have to be used.  For our model, inference is carried out by the
multi-label graph-optimization library
of~\cite{boykov2001,boykov2004,komogorov2004}, using $\alpha$-expansion.
$\alpha$-expansion reduces the multi-label optimization problem to a sequence of
binary optimization problems. Given a labeling \textbf{x}, each pixel $i$ makes
a binary decision to keep its current label or switch to label $\alpha$ ($\alpha
\in \mathcal{L}$).

\begin{figure}[t]
  \centering
  \includegraphics[width=\linewidth]{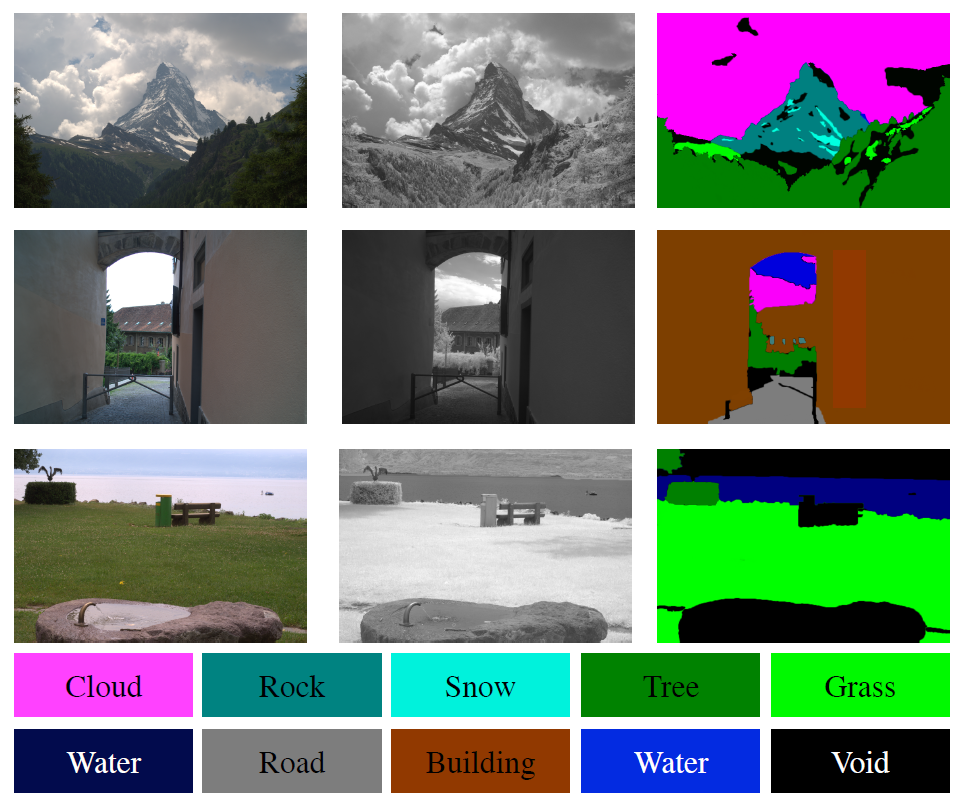}
  \caption{Sample images from our outdoor dataset: RGB (left), NIR
  (middle), groundtruth pixel-level annotation (right).}
  \label{fig:dataset}
\end{figure}

\section{Datasets and experimental setup}
\label{sec:experiment}

\subsection{Proposed datasets}

\begin{figure}[t]
  \centering
  \includegraphics[width=\linewidth]{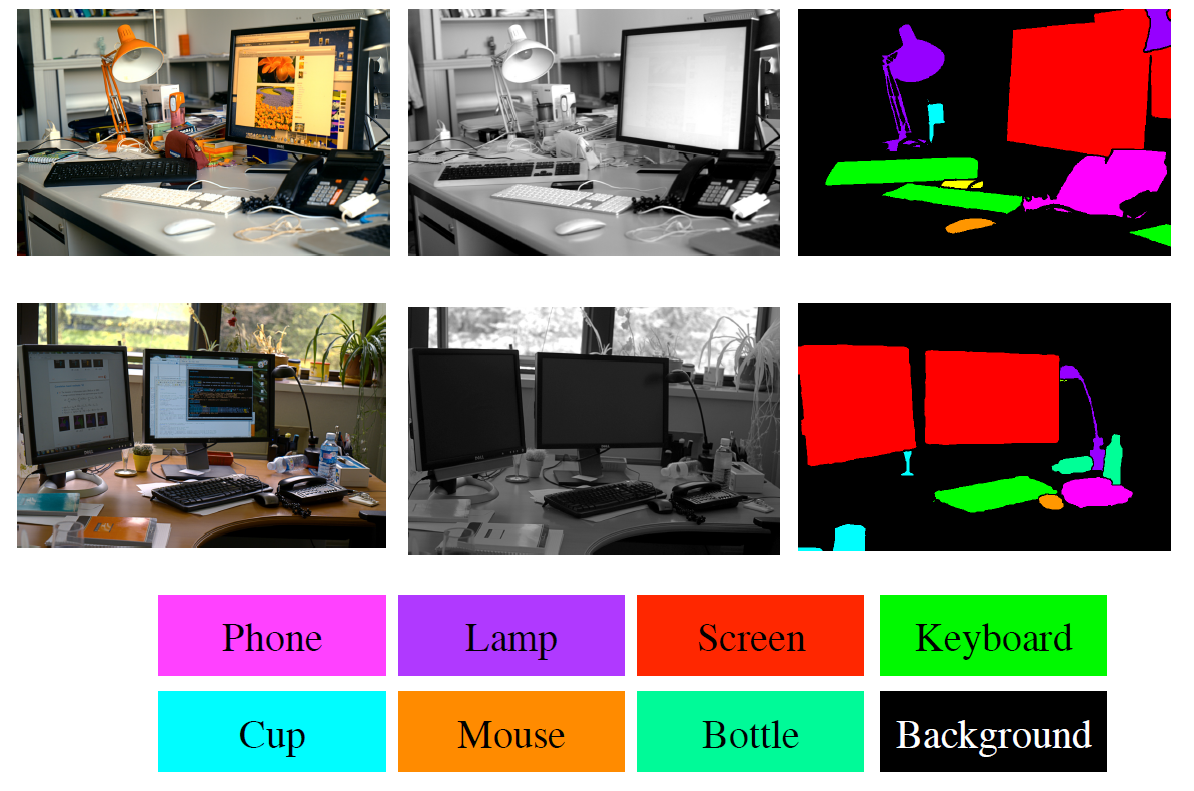}
  \caption{Sample images from our indoor dataset: RGB (left), NIR
  (middle), ground truth pixel-level annotation (right).}
  \label{fig:dataset2}
\end{figure}

In order to evaluate and better understand the gain of incorporating NIR
information in different parts of our CRF model, we labeled a
scene segmentation dataset of 770 images with pixel level annotation considering
 23 semantic classes. Although this dataset
is smaller than the most recent visible only segmentation datasets, it
constitutes the very first test-bed for RGB+NIR semantic segmentation, and
already allows to conduct a deep study, as shown in our experiments.
It is composed of two types of scenes: indoor and outdoor scenes. We conducted
separate experiments on both. Examples of 4-channel images for these two
datasets and their annotations can be seen in Figure~\ref{fig:dataset} and
Figure~\ref{fig:dataset2}.

{\bf The outdoor dataset} was built from the 477 RGB and NIR image pairs
released by~\cite{brown2011}. From these images, we discarded 107 images due to
mis-registration and ambiguity of classes.  The rest of the images were manually
labeled at the pixel level, thus yielding pixel segmentation masks. The labels
were selected from 10 predefined classes\footnote{The number of images that
  contain at least an instance of each
  class in given in brackets.}: {\em building (179), cloud (161), grass (159),
  road (108), rock (80), sky (174), snow (41), soil (78), tree (274), and water
  (79)}. We followed the MSRC dataset's annotation style~\cite{shotton2006}, \ie
each pixel is labeled as belonging to one of the above classes or to a {\em
  void} class. The latter corresponds to pixels whose class is not defined as
part of our classes of interest, or that are too ambiguous to be labeled. Similarly
to~\cite{shotton2006}, in this outdoor dataset, we discarded the pixels of the
{\em void} class from the evaluation.

{\bf The indoor dataset} consists of 400 images that we specifically gathered in
various office environments using sequential capture and registration, as done
in~\cite{brown2011}. The registration between RGB and NIR images was conducted
with the algorithm proposed in~\cite{szeliski2006}. For these images we selected
13 object categories: {\em screen (206), clothing items (184), keyboard (178),
  cellphone (108), mouse (145), office phone (113), cup (163), bottle (130),
  potted plant (77), bag (123), office lamp (70), can (59)}. These objects were
manually segmented and annotated at the pixel level, as in the PASCAL VOC
Challenge~\cite{everingham}, where all pixels not belonging to the predefined
classes are considered as {\em background}. Contrary to the {\em void} class,
the segmentation performance of the predicted background is evaluated.

The {\em background} class, however, is rather diverse. Therefore, instead of
modeling it explicitly, we first predict only the other classes and then we
employ a minimum level of confidence threshold on the predicted classification
scores. If the maximum posterior probability is smaller than a single universal
threshold (in our case $\mathcal{T}=0.5$), the pixel is labeled as {\em
  background}, otherwise it is labeled with the class label for which the
maximum was found. In other words, in our CRF model, $P(X=Background\mid
\textbf{D}) = \mathcal{T}$.

\subsection{Experimental details} 

We extract $32 \times 32$ patches on a regular grid (every 10 pixels), at 5
different scales (the first 5 terms of the geometric series with ratio
$\sqrt{2}$) in a given channel. Hence the coarsest scale corresponds to
re-sizing the patch by a factor of 4 (${\sqrt{2}}^4$) and the finest corresponds
to the original patch (${\sqrt{2}}^0=1$).

Low-level descriptors are computed for each patch. We consider two different
descriptors: SIFT descriptors~\cite{lowe2004} and local color
statistics~\cite{clinchant07}, denoted by $SIFT$ and $COL$ respectively. To
compute these features in any of the considered channels (initial channels
$R,G,B,N$, luma $L$ or alternative color spaces $P_1,P_2,P_3,P_4$), we proceed
as follows.  A single $SIFT$ is 128-dimensional while $COL$ is 32-dimensional,
because we consider the mean and the variance of the color intensity using a 4x4
grid on the patch. Then we concatenate the relevant features, hence $COL_{rgb}$
will be 96 dimensional, $COL_{rgbn}$ will be 128 dimensional, and $SIFT_{rgbn}$
will be 512 dimensional. For a fair comparison, PCA projection is used to reduce
all features to 96 dimensions. In the projected space a GMM-based visual
codebook of 128 Gaussians is built and used to transform each low-level feature
into a Fisher Vector (FV) representation~\cite{Perronnin2010}.  By using the
same PCA projection and the same codebook size, the FV representation of all
descriptors share the same dimension.

Sparse Logistic Regression (SLR) \cite{krishnapuram05} is used for
the classification of the patches and the output is a probability obtained from
$\frac{1}{1+\exp(-s_k)}$, where $s_k,k \in \mathcal{L}$ are the classification
scores of the FVs.  The class probabilities of each pixel $P(X_i=k \mid
\textbf{D})$ are then computed as a weighted average of the patch posteriors as
described in Section~\ref{sec:unary}.

\subsection{Evaluation procedure} 
\label{sec:eval}

In all our experiments, we randomly split the dataset into 5 sets of images (5
folds) and define 5 sets of experiments accordingly. For each experiment, one
fold is used as validation set, one is used as the testing set and the remaining
images are used for training the model. Results (\ie predicted segmentation
maps) for the 5 test-folds are grouped all together and evaluated at once, thus
producing a single score for the entire dataset.

To compare the segmentation results of different methods and parameters, we 
use both region-based and contour-based measures as they were shown to be
complementary~\cite{Csurka2013}. We detail all the measures that we use below.

\textbf{Region-based accuracies} are generally based on the confusion matrix
$\mathbf{C}$ that is computed on an aggregation of predictions (accumulating the predictions) 
for the whole dataset. Sometimes they can be computed at the image-level which
allows to compute statistical significance tests~\cite{Csurka2013}. 
Such confusion matrix is obtained by:
\begin{eqnarray*}
 \mathbf{C}_{kl}= \sum_{I} \lvert \{ p_i\in I | S^I_{gt}(p_i)=k  \,\, \&  \,\, S^I_{pr}(p_i)= l  \} \rvert
\end{eqnarray*}
where $S^I_{gt}$ is the ground-truth segmentation map of the image $I$, 
$S^I_{pr}$  the predicted segmentation map and $\lvert \textbf{.} \rvert$ is 
the number of elements in the set. 
Hence  $ \mathbf{C}_{kl}$ (the element on the row $k$ and column $l$ 
 of matrix $\mathbf{C}$)
represents the number of pixels with ground truth class 
label $k\in \mathcal{L}$, which are predicted with the label $l\in \mathcal{L}$.

Denoting by $\mathbf{G}_k =\sum_{l} \mathbf{C}_{kl}$, 
the total number of ground-truth pixels labeled with $k$, 
and by $\mathbf{P}_l =\sum_{k} \mathbf{C}_{kl}$  
the total number of predicted  pixels labeled with $l$,  
 we can define  the following evaluation measures:
\begin{itemize}
\item {\em Overall Pixel Accuracy ($OA$)}   measures 
the ratio of correctly labeled pixels:
\begin{eqnarray*}
OA=\frac{\sum_{k=l_1}^{l_n}  \mathbf{C}_{kk}} {\sum_{k=l_1}^{l_n} \mathbf{G}_k}
\end{eqnarray*}

\item {\em Per Class Accuracy ($CA$)}  measures the ratio of correctly 
labeled pixels for each class and then averages over all  classes:
\begin{eqnarray*}
CA=\frac{1}{\lvert  \mathcal{L} \rvert} \sum_{k=l_1}^{l_n}  
\frac{\mathbf{C}_{kk}}{\mathbf{G}_k}
\end{eqnarray*}
\item {\em The Jaccard Index ($JI$)}  measures the intersection 
over the union of the labeled segments. This measure is computed by 
dividing the diagonal value $\mathbf{C}_{kk}$ (true positives) by the sum of all 
false positives and all false negatives for a given  class $k\in \mathcal{L}$:
\begin{eqnarray*}
JI=\frac{1}{\lvert  \mathcal{L} \rvert} \sum_{k=l_1}^{l_n} \frac{\mathbf{C}_{kk}}{\mathbf{G}_k+\mathbf{P}_k-\mathbf{C}_{kk}}
\end{eqnarray*}
\end{itemize}
Note that $OA$ and $CA$ correspond to the measures used in general 
to compare  segmentation results on the MSRC dataset~\cite{shotton2006}, whereas 
$JI$  is the measure used in the PASCAL VOC Segmentation 
Challenge~\cite{everingham}.

\begin{table*}[ttt]
\begin{center}
\begin{tabular}{|c||c||c|}
\hline
 Method & Outdoor & Indoor \\
\hline
\begin{tabular}{|c|}
\textbf{Descriptor} \\
\hline
$COL_{rgb}$ \\
$COL_{rgbn}$\\
$COL_{p1234}$\\
\hline
$SIFT_{l}$\\
$SIFT_{n}$\\
$SIFT_{p1}$\\
\hline
$SIFT_{rgb}$\\
$SIFT_{rgbn}$\\
$SIFT_{p1234}$ \\
\hline
\hline
$COL_{rgb} + SIFT_{l}$  \\
$COL_{rgbn} + SIFT_{l}$  \\
$COL_{rgbn} + SIFT_{n}$  \\
$COL_{p1234} + SIFT_{n}$  \\
\end{tabular} &
 \begin{tabular}{|c|c|c|}
 \textbf{CA} & \textbf{OA} & \textbf{JI}\\
\hline
 74.07 & 78.25 & 59.21 \\
 76.18 & 80.56 & 61.94 \\
 76.95 & 80.63 & 63.00\\
\hline
 66.88 & 73.36 & 50.68\\
 67.07 & 73.96 & 51.12\\
 61.01 & 73.44 & 50.91\\
\hline
 75.07 & 80.13 & 60.33\\
 76.47 & 82.38 & 62.41\\
 76.74 & 82.55 & 62.77 \\
\hline
\hline
 79.17 & 83.52 & 65.85\\
80.18 & 84.76 & 67.34 \\
 80.13 & 84.88 & 67.40\\
 \textbf{80.91} & \textbf{85.19} & \textbf{68.46}\\
\end{tabular} & 
 \begin{tabular}{|c|c|c|}
 \textbf{CA} & \textbf{OA} & \textbf{JI}\\
\hline
 39.94 & 50.49 & 24.23\\
 45.33 & 56.03 & 28.74\\
 44.57 & 54.27 & 28.10\\
\hline
 49.19 & 47.55 & 32.49\\
 48.32 & 43.46 & 31.54\\
 48.30 & 44.02 & 31.22\\
\hline
 49.75 & 51.91 & 33.03\\
 \textbf{53.79} & 58.98 & \textbf{36.77}\\
 49.75  & 57.41 & 33.09\\
\hline
\hline
 49.47 & 56.50 & 32.36\\
 53.64 & 60.95  & 36.26\\
 53.20 & \textbf{61.13}  & 35.78\\
 52.78 & 60.23 & 35.44\\
\end{tabular}\\
\hline
\end{tabular}
\caption{Average of per-class accuracies, overall accuracies, and Jaccard index of the
  segmentation results obtained for different local descriptors and their
  combinations, on the outdoor and indoor datasets.\label{table:desccomp}}
\end{center}
\end{table*}

\textbf{The trimap accuracy (TrimapAcc)} evaluates 
segmentation accuracy around boundaries~\cite{kohli2009}. 
The idea of this measure is to build a narrow band around 
each contour and to compute pixel accuracies $OA$, thus evaluating only the 
pixels within the given band. As a single band gives only partial evaluation, 
the size of this band $r$ is varied and the 
overall accuracy values (denoted here by $T(r)$) is plotted as a curve.

\textbf{Tests of statistical significance.} To examine if our results are
statistically different, we also compute a paired t-test on the results per
image.  In comparing the classification rate of two strategies A and B, the null
hypothesis $H_0$ is that the results obtained by A and B are independent random
samples from a normal distributions with an equal mean. The paired t-test
computes the probability, $p$-value, of the hypothesis $H_0$. A typical value
for rejecting the hypothesis that the scores for different representations come
from the same population is $p-\text{value} < 0.05$.  In such cases, we can say
that the two methods generating the respective mean results are significantly
different.

\section{Experimental results}
\label{sec:results}

In Section~\ref{sec:model}, we have described different ways to integrate 
NIR information into our segmentation framework. In this section, we 
investigate and compare these different options through a set of experiments. 
First, we consider only the recognition part (\ie only the unary term) of our
model and compare  
different descriptors and combinations of visible and NIR based features. The 
study for the regularization part (adding the pairwise energy term) is conducted only for 
the best performing recognition models. 

\subsection{Recognition component}

To compare the recognition ability of the different descriptors, we produce a semantic segmentation by 
assigning pixels to their most likely label with
\begin{eqnarray*}
\textbf{x}^*  =   \textrm{argmax}_{\textbf{x} \in \mathcal{L}^C} \,
\sum_{i \in \nu} P(X_i=\textbf{x}_i \mid \textbf{D})
\end{eqnarray*}
given the observation \textbf{D}. This is equivalent to the full model Eq (\ref{eq.fullcrf}) when 
 $\lambda = 0$. 

For each pixel, the label corresponding to the highest score is retained,
yielding a predicted segmentation map.  In the case of the indoor dataset, this
score is further compared to the threshold $\mathcal{T}$ (in our case we used
0.5) and if the highest score is below this threshold, the pixel is assigned to
the {\em background} class.

The accuracy of the predicted segmentations is then evaluated with the different
region-based accuracy measures described in Section~\ref{sec:eval}. Note that
for these experiments there is no regularization term enforcing the region
boundaries to follow image contours, so we do not use contour based evaluation
measures.

In Table~\ref{table:desccomp} (upper part), we show the segmentation results
obtained with region-based accuracy measures for different local descriptors.
From these tables we can observe that the accuracy of the recognition using
$COL$ features is significantly higher when NIR descriptors are also considered
with those of the RGB image ($COL_{rgbn}$, $COL_{p1234}$), compared to the
visible only scenario ($COL_{rgb}$). Similarly, combining visible and NIR
features, $SIFT_{rgbn}$ and $SIFT_{p1234}$ outperform $SIFT_{rgb}$. 
%
$SIFT_{n}$ performs similarly to $SIFT_{l}$, leading to slightly better
performance in outdoor environments, and slightly worse for indoor ones.
The reason might be that, in the NIR image, material-intrinsic texture
properties are captured, which might be insufficient to describe the appearance
of our objects in the indoor dataset. For most classes in the outdoor dataset,
however, this appearance seems to be better captured in NIR than in the L
channel.

In both cases, the best single descriptor results are obtained with
multi-spectral SIFT, when both the visible and NIR images are considered. Note
that these features incorporate both texture (explicitly) but also color
(implicitly, considering the SIFT in multiple channels).
Another way to combine color and texture is by early or late fusion of $COL$ and
$SIFT$. As \cite{salamati2011} clearly showed that the late fusion of these
features outperforms early fusion for image categorization; here we do not
consider the latter. The results of late fusion between different $COL$ and
$SIFT$ features are shown in the lower part of Table~\ref{table:desccomp}. Note
that $COL_{rgb} + SIFT_{l}$ corresponds to our visible baseline for the
recognition part\footnote{Note that it is similar to the approach of
  \cite{csurka2011} without region labeling and without global score-based fast
  rejection.}.

\begin{table*}[ttt]
\begin{center}
\begin{tabular}{|c||c||c|}
\hline
 Method & Outdoor & Indoor \\
\hline
\begin{tabular}{|c|c|}
\textbf{Descriptor} & \textbf{Pairwise} \\
\hline
\multirow{3}{*}{$SIFT_{rgb}$} & $VIS$  \\
 & $NIR$ \\ 
 & $VIS+NIR$ \\ 
\hline
\multirow{3}{*}{$SIFT_{rgbn}$} & $VIS$  \\
 & $NIR$ \\ 
 & $VIS+NIR$ \\ 
\hline
 \multirow{3}{*}{$SIFT_{p1234}$} & $VIS$ \\
 & $NIR$ \\ 
 & $VIS+NIR$ \\
\hline
\hline
\multirow{3}{*}{$COL_{rgb}+SIFT_{l}$} & $VIS$  \\
 & $NIR$ \\ 
 & $VIS+NIR$ \\ 
\hline
\multirow{3}{*}{$COL_{rgbn}+SIFT_{l}$} & $VIS$ \\
 & $NIR$ \\ 
 & $VIS+NIR$ \\
\hline
\multirow{3}{*}{$COL_{rgbn}+SIFT_{n}$} & $VIS$ \\
 & $NIR$ \\ 
 & $VIS+NIR$ \\
\hline
  \multirow{3}{*}{$COL_{p1234}+SIFT_{n}$} & $VIS$ \\
 & $NIR$ \\
 & $VIS+NIR$ \\
\end{tabular} &
\begin{tabular}{|c|c|c|}
 \textbf{CA}  & \textbf{OA} & \textbf{JI}\\
\hline
77.02 & 82.40  & 62.90 \\
77.00 & 82.15 & 62.86\\
77.05 & 82.40 & 62.91  \\
\hline
78.07 & 84.02 & 64.54 \\
78.07 & 83.99 & 64.54 \\
78.25 & 84.17 &  64.80 \\
\hline
 78.30  & 84.22 &  67.97 \\
78.30  & 84.11 & 67.97 \\ 
 78.35  & 84.27 & 68.04 \\
\hline
\hline
 79.97  & 84.56 & 67.14 \\
 80.05  & 84.73 & 67.06\\
  80.24 & 84.87 &   67.40\\ 
\hline
81.22 & 85.90 & 68.82 \\
81.15 & 85.88 & 68.78 \\
81.22 & 85.97 & 68.87 \\
\hline
81.14 & 86.08 & 68.89 \\
 81.20 &86.07  & 69.02 \\ 
 81.31 & 86.22 & 69.15 \\
\hline
81.69 & 86.22 & 69.61 \\
81.56  & 86.01 & 69.47 \\ 
\textbf{81.86}  &  \textbf{86.34} &  \textbf{69.86} \\
\end{tabular}
&
\begin{tabular}{|c|c|c|}
 \textbf{CA}  & \textbf{OA} & \textbf{JI}\\
\hline
51.60 & 60.94 & 34.69\\
 51.47 & 60.98 & 33.42\\ 
51.75& 60.86 & 34.79\\ 
\hline
 55.3 & 68.0 & 38.5\\
 55.16 & 68.0 & 38.44\\ 
\textbf{55.67} & 68.02 & \textbf{38.86}\\
\hline
50.72 & 65.82 & 34.10\\
 51.27 & 66.00 & 34.67\\ 
51.26 & 65.85 & 34.59\\
\hline
\hline
50.84 & 63.58 & 33.61 \\
50.70 & 63.81 & 33.46 \\
51.113 & 63.57 & 33.87  \\
\hline
 54.54 & 68.23 & 37.06\\
 54.63 & 68.24 & 37.15\\ 
 54.65& 68.27 & 37.11\\
\hline
 53.86 & 68.78 & 36.37\\
 53.97 & 68.75 & 36.48\\ 
54.37 & \textbf{68.78} & 36.89\\
\hline
53.54 & 67.97 & 36.32\\
 53.77 & 67.93 & 36.05\\ 
 53.91  & 67.87 & 36.37\\
\end{tabular} 
\\
\hline
\end{tabular}
\caption{Results for the full CRF model for the outdoor and the indoor datasets
  using regularization based on the RGB image ($VIS$),  on the  NIR information
($NIR$) and on both ($VIS + NIR$).\label{table:graph}}
\end{center}
\end{table*}

Comparing the results of late fusion of $COL$ and $SIFT$ to the multi-spectral
SIFT, we can observe the following: In the case of the outdoor dataset, the late
fusion of $COL$ and $SIFT$ clearly outperforms the multi-spectral SIFT, whereas
this is not true in the case of the indoor dataset where the two strategies
yield similar results.  The main reason could be that color (RGB) in the case of
outdoor scenes is much more important than in indoor scenes where most objects
have different colors that are not specific to a given class. This observation
is confirmed by the low performance of $COL_{rgb}$ compared to the $SIFT_{l}$ in
the case of indoor dataset; whereas for the outdoor dataset $COL_{rgb}$
significantly outperforms $SIFT_{l}$, thus showing how important the color is
for predicting the appearance of these scene classes (\eg \textit{sky},
\textit{grass}, \textit{snow}, etc.). 


Although the best strategy seems to be scene type dependent, $COL_{rgbn} +
SIFT_{n}$ yields close to best results in both cases.

\subsection{Full CRF model}

For these experiments, we consider the most promising recognition models, both for
visible only and for the visible+NIR images, and we apply the full CRF model with 
regularization based on  the RGB image ($VIS$),  on the  NIR information
($NIR$) and on both ($VIS + NIR$). We used a fixed weight parameter 
$\lambda=5$ for the regularization part (see Eq (\ref{eq.fullcrf})) in all our experiments.

\begin{figure}[t]
\begin{center}
\includegraphics[width=0.5\textwidth]{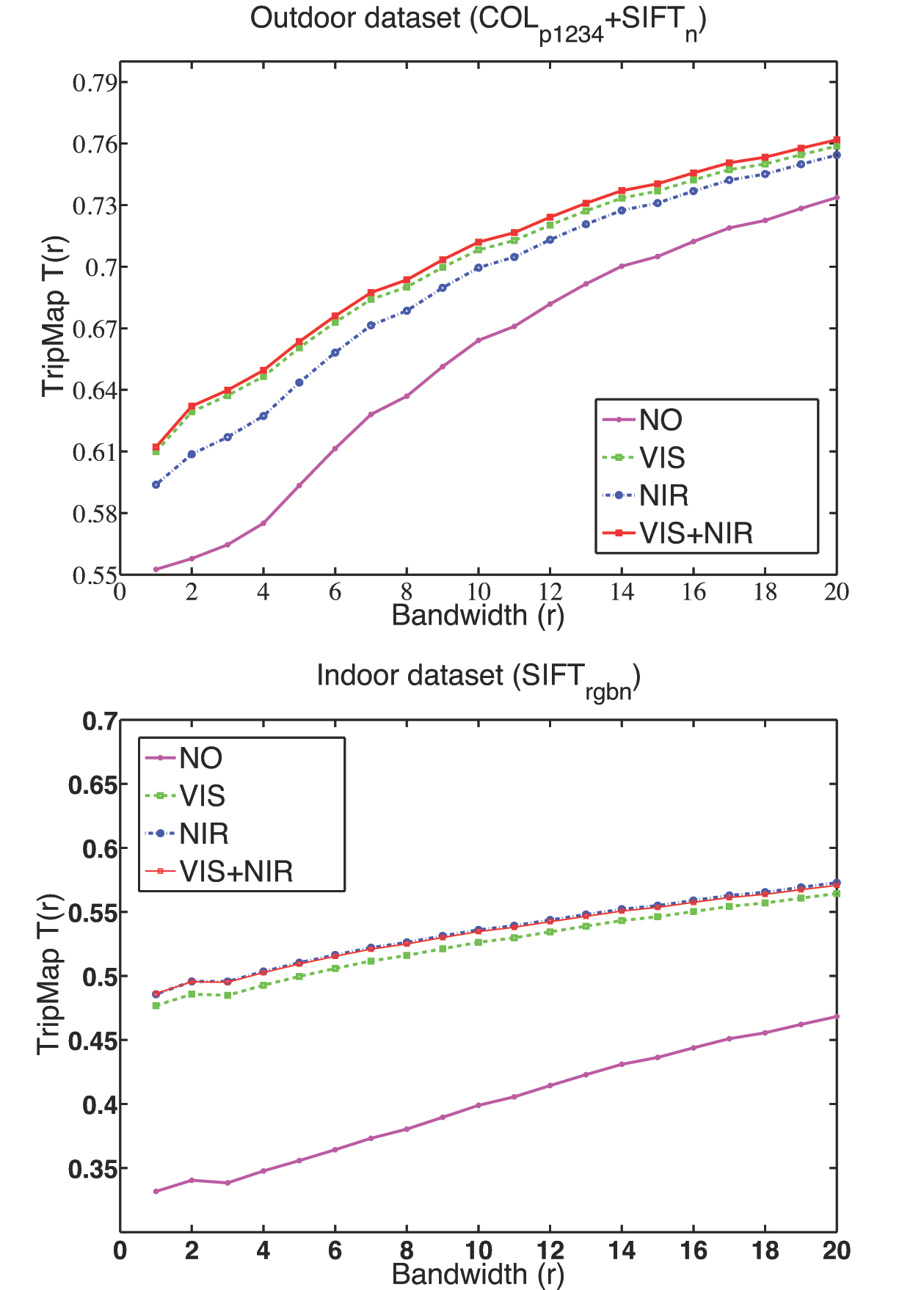} 
\caption{The TrimapAcc plots with different pairwise potentials using 
 $COL_{p1234} + SIFT_n$  (top- for the outdoor dataset) 
 respectively $SIFT_{rgbn}$  (bottom- for the indoor dataset) as unary potential.}\label{fig:trimap}
\end{center}
\end{figure}

\begin{figure}[t]
\begin{center}
\includegraphics[width=0.5\textwidth]{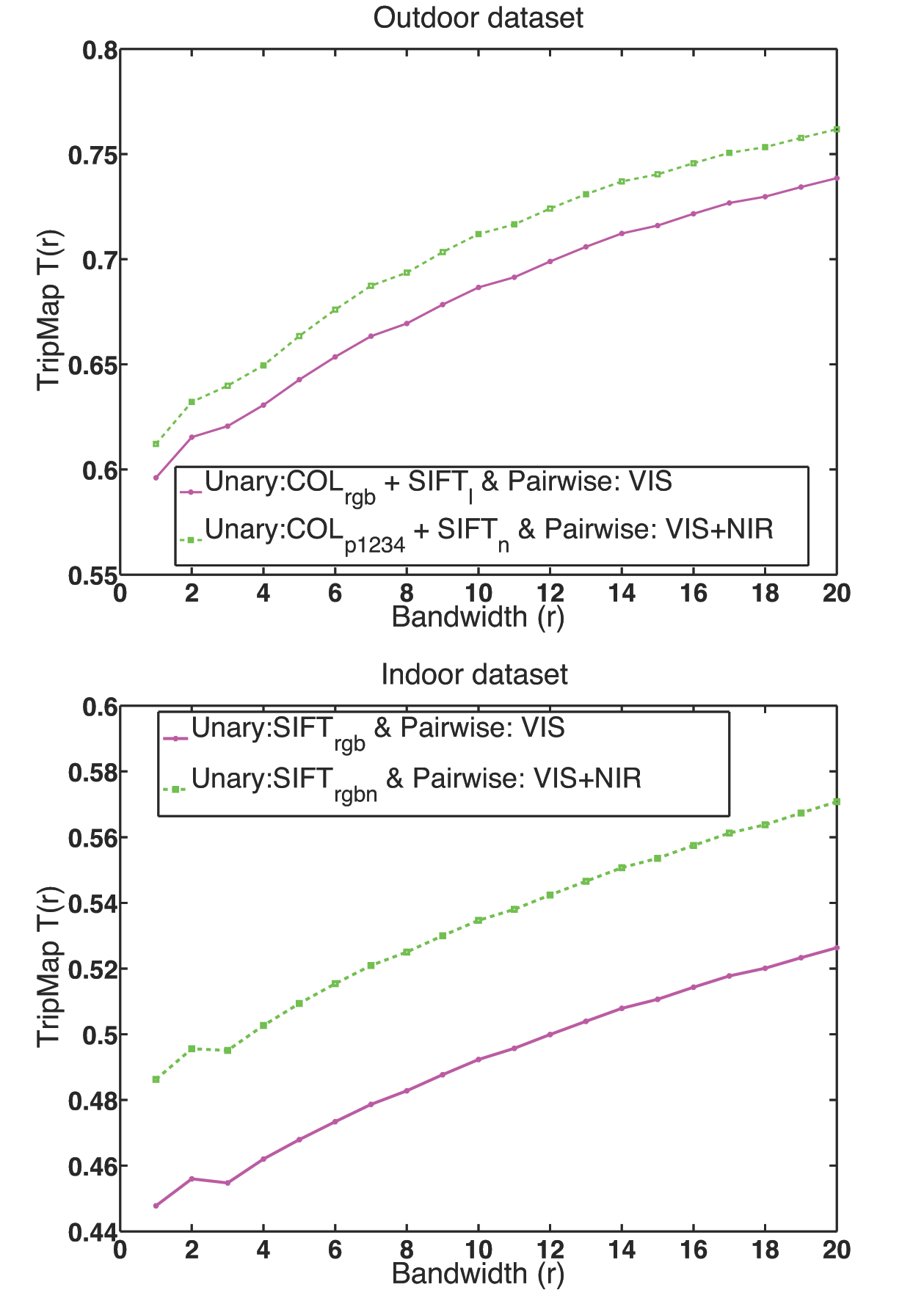} 
\caption{The TrimapAcc plots compare the border accuracy of the results of the
  visible only scenario and the proposed strategy, top- for the outdoor dataset
  and bottom- for the indoor dataset.}\label{fig:trimap2}
\end{center}
\end{figure}

Results are shown in Table~\ref{table:graph}.
First, we can see that for all the descriptors, 
the quality of semantic segmentation consistently increases when we consider 
both visible and NIR channels ($VIS+NIR$)  in the pairwise 
potential, compared to using only visible or only NIR. Second, 
as expected, the regularization term (from any of the three strategies) improves the 
segmentation results obtained for any recognition model 
compared to the segmentation without regularization ($\lambda=0$, presented in
Table~\ref{table:desccomp}). 
Finally, the ranking between the models with 
different  descriptors  is similar with or without regularization, 
given a regularization model (\eg $VIS$ or $VIS+NIR$).
This is again not surprising, as we use the same regularization term 
that is independent of the feature used in the recognition part. 
Again $COL_{rgbn}+SIFT_{n}$ and $COL_{p1234}+SIFT_{n}$ lead to 
the best performances in the case of the outdoor dataset, and $SIFT_{rgbn}$
performs best in the case of the indoor dataset, $COL_{rgbn}+SIFT_{n}$ 
being second best. Compared to the visible-only baselines 
$COL_{rgb}+SIFT_{l}$ or $SIFT_{rgb}$ with visible image-based regularization
($VIS$), they are significantly better and statistically different 
at the 95\% confidence level according to the paired t-test applied to 
the score distributions. 

For these experiments, we are mainly interested in the gain we obtain when
comparing the $VIS$ or $NIR$-based regularization with the $VIS+NIR$-based
regularization. Considering region-based evaluation measures, we can see only
slight improvements even if the paired t-test often shows significant
differences between the corresponding score distributions. Therefore, to better
evaluate the gain obtained by combining the $NIR$ and $VIS$ channels in the
regularization, we also evaluate some of these results with contour based
measures.

We apply the trimap accuracy evaluation of~\cite{kohli2009} (overall pixel
accuracy in the neighborhood of object boundaries) and we show the results as a
function of the boundary size in Figures~\ref{fig:trimap} and \ref{fig:trimap2}.
From these results, we can first notice the importance of the regularization. In
both cases, any of the potentials that we use as regularizer leads to a
significant improvement (statistically different at the 95\% confidence level
according to the paired t-test) on the results obtained by recognition alone
($NO$, for no pairwise). Comparing different edge potentials, we can see that,
in outdoor scenes, the visible image leads to better segmentation than NIR image
alone\footnote{This behavior can be partially explained by the fact that the
  manual annotation was done in the RGB images and in some images of the outdoor
  dataset, the position of some objects, such as clouds or cars, are different
  between the two representations because visible and NIR images were acquired
  in two consecutive shots. In the indoor scenes there is no movement between
  the two shots, as no moving objects were present and this bias is not
  observed.}.  Fixing the band $r$ at 5 pixels and running the t-test, we found
that all $T(5)$ results are statistically different at a 95\% confidence level.

In order to show also some qualitative comparisons, Figure~\ref{fig:vis} shows a
few segmentation results obtained with the visible baseline and the best visible
+ NIR setting. A visual inspection of these results allows to observe the
positive qualitative influence of incorporating NIR information in the
segmentation model.

\begin{figure}[t]
\begin{center}
\begin{tabular}{cccc}
\hspace{-0.3cm}\includegraphics[height=5.6cm]{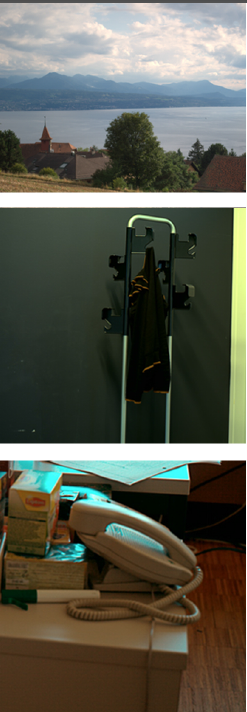}&
\hspace{-0.2cm}\includegraphics[height=5.6cm]{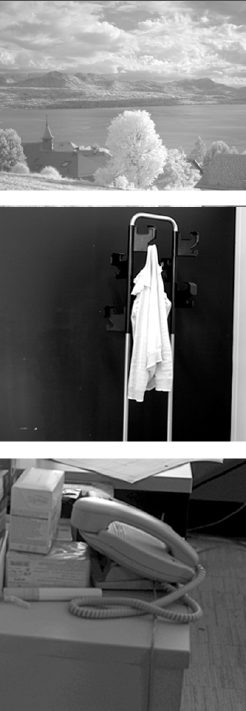}&
\hspace{-0.3cm}\includegraphics[height=5.6cm]{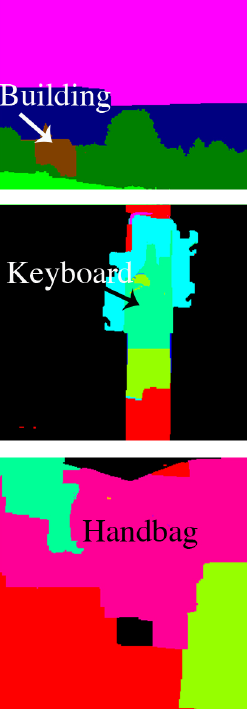}&
\hspace{-0.3cm}\includegraphics[height=5.6cm]{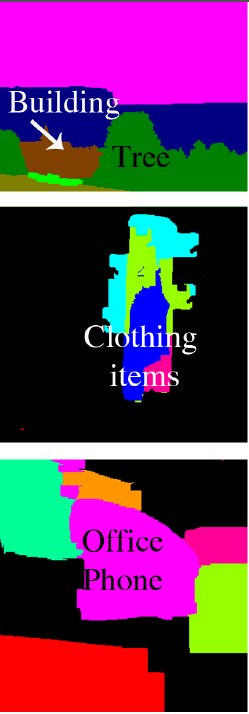}\\
\hspace{-0.3cm} \tiny RGB & \hspace{-0.3cm} \tiny NIR & \hspace{-0.6cm} \tiny 
RGB-only results & \hspace{-0.5cm} \tiny  RGB+NIR results
\end{tabular}
\caption{Examples from both outdoor and indoor datasets. Note that the material
  dependency of NIR images results helps both in recognition and a 
 more accurate detection of object boundaries.}\label{fig:bordsamp}
\end{center}
\end{figure}

\section{Class-based analyses and discussion}
\label{sec:discussion}

This section takes a deep dive into some of the results presented in the previous
section in order to visualize, analyze and compare class-by-class results.  We
focus our analysis on the segmentation results obtained with the best visible
baseline ($COL_{rgb}+SIFT_{l}$ for the outdoor dataset and $SIFT_{rgb}$ for the
indoor dataset, with visible-only pairwise) and the best RGB+NIR integrated
strategy ($COL_{p1234}+SIFT_{n}$ for the outdoor dataset and $SIFT_{rgbn}$ for the
indoor dataset, with VIS + NIR pairwise).

\begin{figure*}[t]
\begin{center}
   \includegraphics[width=13cm]{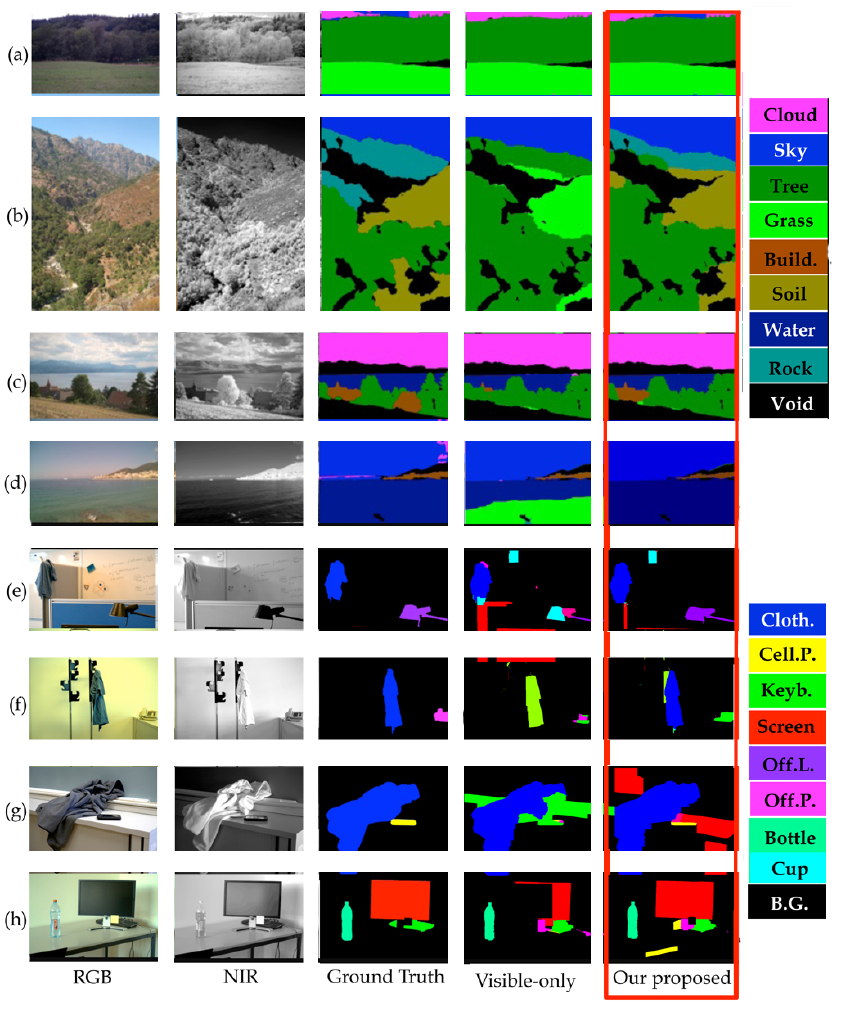}
\end{center}
\caption{Sample segmentation results for the outdoor and the indoor datasets.}
\label{fig:vis} 
\end{figure*}

We first show confusion matrices between the classes in Table~\ref{table:visconmat} and
Table~\ref{table:visconmat2}. We also show qualitative results in
 Figure~\ref{fig:bordsamp},
Figure~\ref{fig:watersamp}, Figure~\ref{fig:hazesamp},
Figure~\ref{fig:screencloth}, Figure~\ref{fig:clothres},
Figure~\ref{fig:flowerpot}, Figure~\ref{fig:skygrass}
and Figure~\ref{fig:vis}, that we all discuss in detail below.

In general, in both datasets, we observe that borders are more precisely
detected when NIR information is incorporated in the pairwise potential. This
can be explained by the material dependency of NIR responses that might reduce
the influence of wrong edges due to clutter, or might result in more contrasted
edges between classes. This information, used in the regularization part of our
model, helps better aligning borders between regions with the material change,
as we can see in the examples of the Figure~\ref{fig:bordsamp}. We now analyze
in depth the different classes, and discuss the datasets individually.

\subsection{Outdoor scenes}

First we discuss the results obtained for the outdoor dataset. This dataset
contains mostly what is usually referred as ``stuff'', \eg background classes
that are difficult to count. Based on Table~\ref{table:visconmat} results, we
can make the following observations.

\textbf{Natural materials.}  First we discuss the relations between {\em tree,
  grass} and {\em soil}.  Looking at the yellow boxes in
Table~\ref{table:visconmat}, we observe the following. {\em Tree} and {\em
  grass} are more confused in NIR because they consist of the same material and
have roughly the same texture. {\em Grass} and {\em soil} are less confused in
NIR. The reason could be that the pigment in vegetation (Chlorophyll) is
very reflective in the NIR part of the spectrum, whereas {\em soil} has a lower
reflectance in this part that makes it appear very differently in the NIR
representation of the scene.  {\em Soil} and {\em tree} get more confused when
NIR information is incorporated into our model. This can be explained by the
fact that the wooden part of the {\em tree} has almost the same pixel values as
{\em soil}, thus incorporating the $COL$ descriptor in the NIR part of the
spectrum could be a source of confusion.

\begin{table*}[t]
\centering
\begin{tabular}{c}
\includegraphics[width=\textwidth]{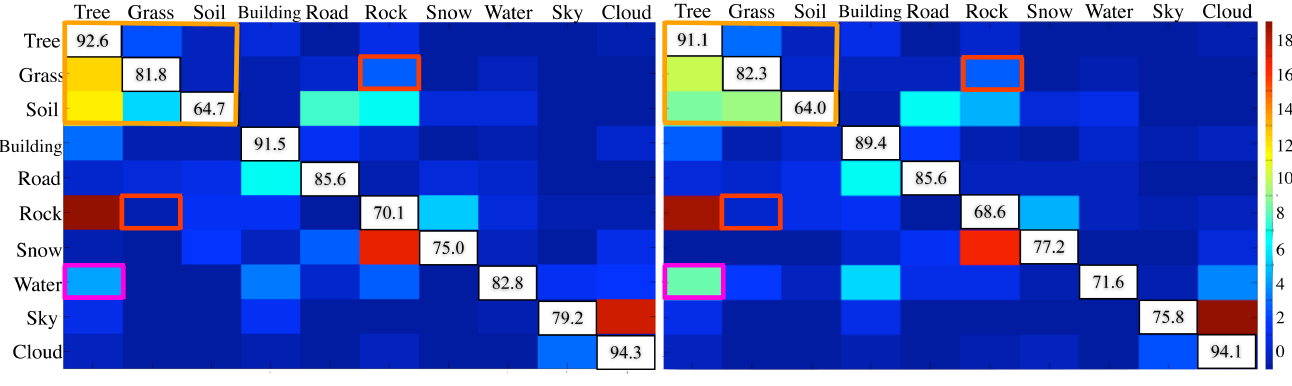} \\
\end{tabular}
\caption{Outdoor dataset. Left, confusion matrix of $COL_{p1234}+SIFT_{n}$ and
  visible+NIR pairwise. Right, confusion matrix for the best visible scenario
 ( $COL_{rgb}+SIFT_{l}$ with visible-only pairwise). Values on the diagonal of the confusion matrices
 correspond to the individual class accuracies.}
\label{table:visconmat}
\end{table*}

{\em Rock} and {\em grass} get more confused in the absence of NIR information
(see the red boxes in Table~\ref{table:visconmat}). Chlorophyll in {\em grass}
is very reflective and discriminative in NIR images and {\em rock} is relatively
more absorbent. Due to such material dependency of NIR images, incorporating
this information into our model decreases the confusion of these two classes.

\begin{figure}[t]
\begin{center}
\begin{tabular}{cccc}
\hspace{-0.3cm}\includegraphics[height=2.2cm]{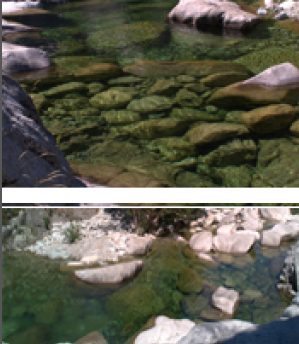}&
\hspace{-0.2cm}\includegraphics[height=2.2cm]{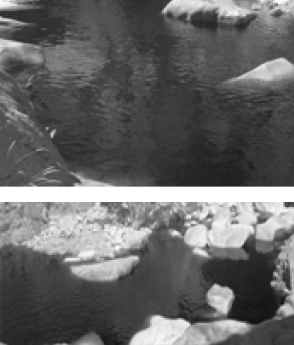}&
\hspace{-0.2cm}\includegraphics[height=2.2cm]{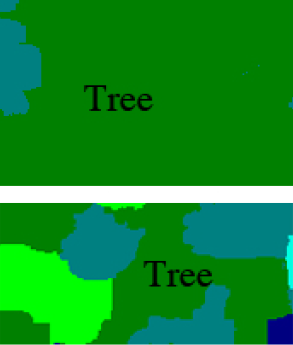}&
\hspace{-0.3cm}\includegraphics[height=2.2cm]{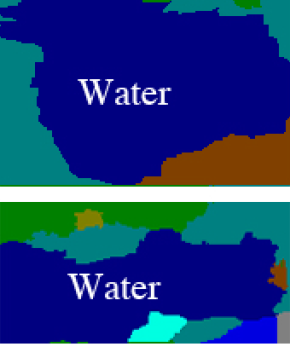}\\
\hspace{-0.3cm} \tiny RGB & \hspace{-0.2cm} \tiny NIR & \hspace{-0.6cm} \tiny 
$COL_{rgb}+SIFT_l$ & \hspace{-0.5cm} \tiny  $COL_{pc1234}+SIFT_n$
\end{tabular}
\caption{In the visible only scenario, the class {\em water} is often confused
  with {\em tree}. The main reason is that the reflection of close-by trees or
  the presence of vegetation in the water make the color of the water very
  similar to the one from trees.}\label{fig:watersamp}
\end{center}
\end{figure}

\textbf{{\em Water.}} \textit{Water} has a mirror reflection in RGB images, so the reflection of
close-by trees in {\em water} makes the $COL_{rgb}$ for the water class very
close to $COL_{rgb}$ for {\em tree}. In NIR images, however, {\em water} is very
absorbent and appears very dark, hence it is confused less with other classes of
material.  See Figure~\ref{fig:watersamp} for an illustration. In other words, even
if in the RGB image, due to the reflection, the color or even the texture of the
related region in the water lead to confusion with the reflected class ({\em
  tree}, {\em sky}, far away mountains), \textit{water} has a unique appearance in the
NIR leading to significantly improved recognition, compared to visible baseline
(see the magenta boxes in Table~\ref{table:visconmat} and the class accuracy of
the water class on the diagonal).

\begin{table*}[t]
\centering
\begin{tabular}{c}
\includegraphics[width=\textwidth]{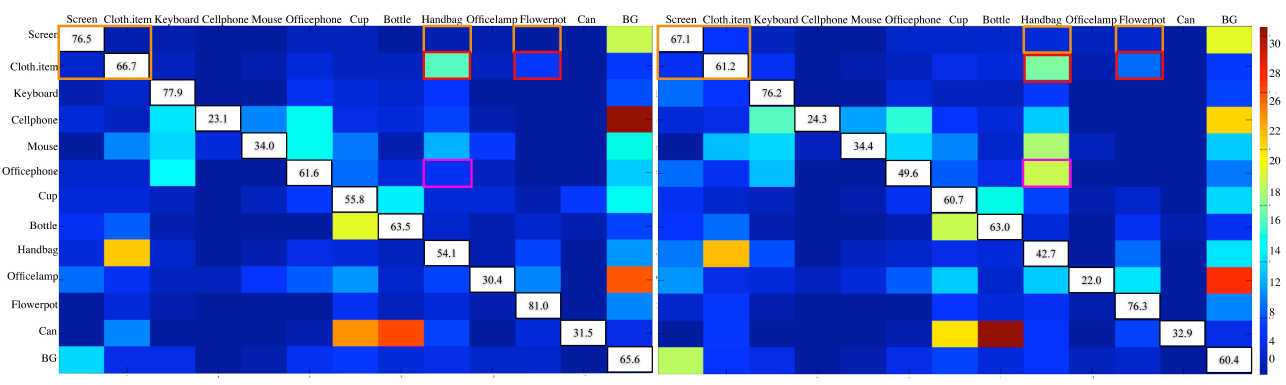}\\
\end{tabular}
\caption{Indoor dataset. Left, confusion matrix of $SIFT_{rgbn}$ and visible+NIR
  pairwise. Right, confusion matrix for the best visible scenario ( $SIFT_{rgb}$
  with visible-only pairwise). Values on the diagonal of the confusion matrices
  correspond to the individual class accuracies.}
\label{table:visconmat2}
\end{table*}

\begin{figure}[t]
\begin{center}
\begin{tabular}{cccc}
\hspace{-0.3cm}\includegraphics[height=1.8cm]{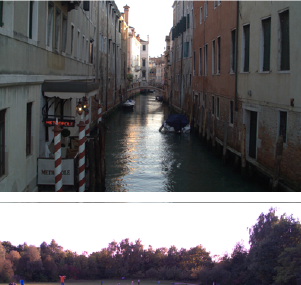}&
\hspace{-0.3cm}\includegraphics[height=1.8cm]{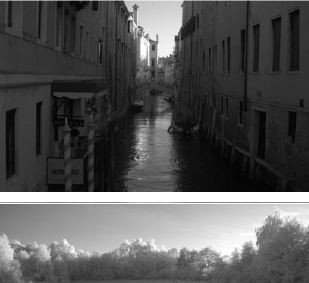}&
\hspace{-0.3cm}\includegraphics[height=1.8cm]{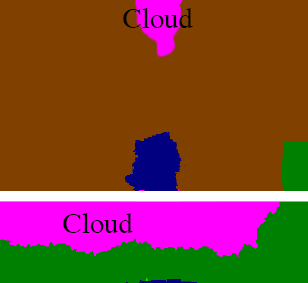}&
\hspace{-0.3cm}\includegraphics[height=1.8cm]{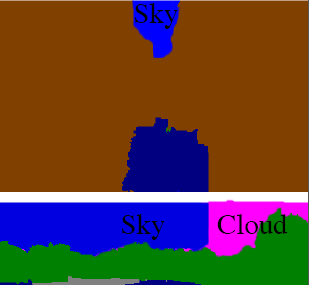}\\
\hspace{-0.3cm} \tiny RGB & \hspace{-0.3cm} \tiny NIR & \hspace{-0.6cm} \tiny 
$COL_{rgb}+SIFT_l$ & \hspace{-0.5cm} \tiny  $COL_{pc1234}+SIFT_n$
\end{tabular}
\caption{Examples from the outdoor dataset. Note the better classification and
  recognition of $clouds$ and $sky$ when NIR information is
  incorporated. }\label{fig:hazesamp}
\end{center}
\end{figure}

\textbf{Haze.} Finally, the benefit of using the NIR channel in the presence of
haze can be observed particularly in the case of {\em sky}, {\em tree} and {\em
  rock} classes, the latter often representing mountains. As stated by
Rayleigh's law, the intensity of light scattered from very small particles
($<\lambda/10$) is inversely proportional to the fourth power of the wavelength
$\lambda$ (\ie $\propto 1/\lambda^4$)~\cite{fredembach2008}. Particles in the
air (haze) satisfy this condition and are therefore scattering more in the
short-wavelength range of the spectrum. Thus, when images are captured in the
NIR part of the spectrum, atmospheric haze is less visible and the sky becomes
darker (see Figure~\ref{fig:hazesamp}). The ``haze transparency'' characteristic
of NIR images results in sharper images for distant objects.  In particular,
vegetation at a distance in the visible image is smoothed and bluish, which can
affect the performance of texture and color features in the classification
task. The sharp and haze-free appearance of vegetation in NIR images helps with
classification and leads to better segmentation (see also the diagonal in
Table~\ref{table:visconmat}).

\subsection{Indoor scenes}

The indoor dataset mostly contains categories that are related to object
classes, also referred as ``things''.  From the results of
Table~\ref{table:visconmat2} we can make the following observations.

\begin{figure}[t]
\begin{center}
\begin{tabular}{cccc}
\hspace{-0.3cm}\includegraphics[height=2.2cm]{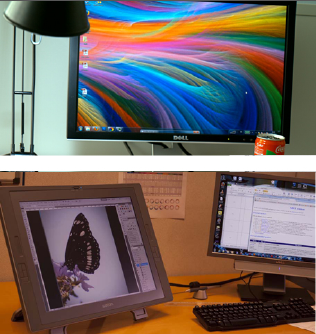}&
\hspace{-0.3cm}\includegraphics[height=2.2cm]{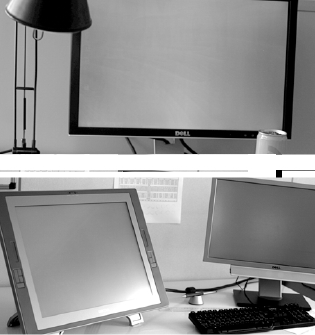}&
\hspace{-0.3cm}\includegraphics[height=2.2cm]{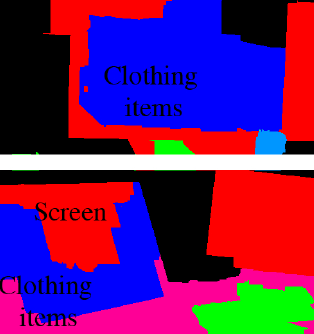}&
\hspace{-0.3cm}\includegraphics[height=2.2cm]{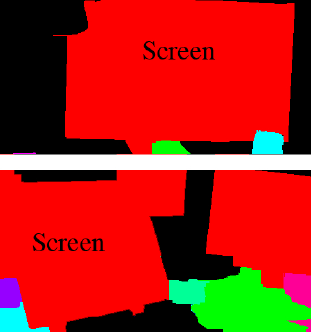}\\
\hspace{-0.3cm} \tiny RGB & \hspace{-0.3cm} \tiny NIR & \hspace{-0.6cm} \tiny 
$SIFT_{rgb}$ & \hspace{-0.5cm} \tiny  $SIFT_{rgbn}$
\end{tabular}
\caption{Multi-spectral-SIFT ($SIFT_{rgbn}$) outperforms the best visible-only
  scenario in recognition of colorful classes where the material properties of
  the classes are different.}\label{fig:screencloth}
\end{center}
\end{figure}

\textbf{Color distractors.} The class {\em screen} is confused with colorful
classes such as {\em clothing item}, {\em handbag}, and {\em flowerpot}, in the
absence of NIR information (see the orange boxes in
Table~\ref{table:visconmat2}). Visible images of class {\em screen} contain many
colorful patches (presence of colorful pictures on the screen), however, the
appearance of this class is consistently the same in NIR images.  As we can see
in Figure~\ref{fig:screencloth} the content displayed on the screen is not
visible anymore in the NIR image and hence incorporating this information yields
to better recognition accuracy.

\begin{figure}[t]
\begin{center}
\begin{tabular}{cccc}
\hspace{-0.3cm}\includegraphics[height=3.2cm]{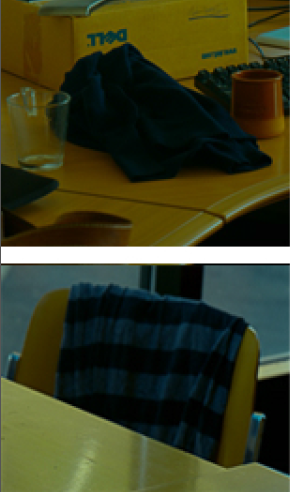}&
\hspace{-0.3cm}\includegraphics[height=3.2cm]{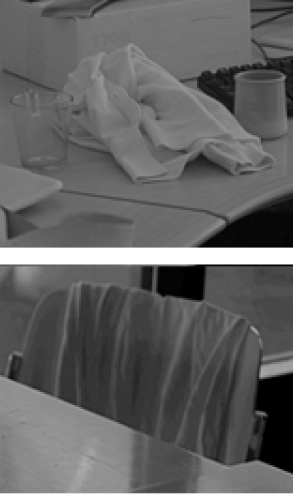}&
\hspace{-0.3cm}\includegraphics[height=3.2cm]{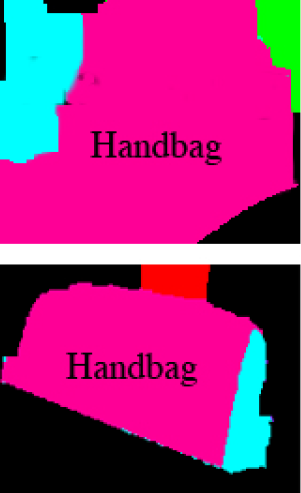}&
\hspace{-0.3cm}\includegraphics[height=3.2cm]{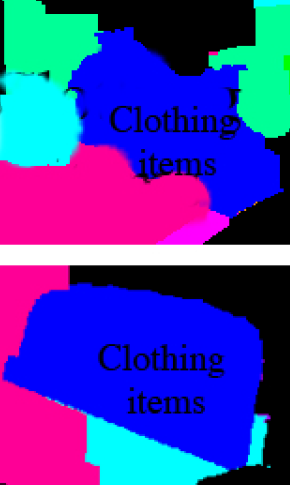}\\
\hspace{-0.3cm} \tiny RGB & \hspace{-0.3cm} \tiny NIR & \hspace{-0.6cm} \tiny 
$SIFT_{rgb}$ & \hspace{-0.5cm} \tiny  $SIFT_{rgbn}$
\end{tabular}
\caption{Due to very different material characteristics of {\em Clothing Item}
  and {\em Handbag}, as well as material dependency of NIR information, the
  confusion of such classes is significantly less in presence of NIR information.}
\label{fig:clothres}
\end{center}
\end{figure}

\textbf{Fabric.} {\em Clothing item} is confused with {\em handbag} in the
visible-only scenario (see the red boxes in Table~\ref{table:visconmat2}),
mainly because such classes appear to look very similar (both in color values
and texture measures). As these two classes are mostly made of different
material ({\em clothing item} is mostly made of natural fabrics such as cotton
or wool, or synthetic fabrics such as polyester or nylon, whereas {\em handbag}
is made of suede or artificial suede, which are types of leather with a
napped finish), they look significantly different in NIR images. See
Figure~\ref{fig:clothres} for an illustration.

\textbf{Man-made objects} In the proposed scenario, {\em cellphone} is generally
confused with the {\em background} class. In the visible only scenario, this class
is more confused with {\em keyboard} as well as {\em background}. The low
performance of our framework for this class is mostly due to the fact
that the number of samples in this class is very low, so the classifier did not
have enough training samples to learn from. A fewer number of training samples
is also a factor for poor and unreliable performance for the {\em cup}, {\em can},
and {\em bottle} classes.

Surprisingly, the {\em Office-phone} class is significantly confused with the class {\em
  handbag} in the visible only scenario. However, when we add information from the
NIR channel, this confusion significantly drops as shown in the red boxes in
Table~\ref{table:visconmat2} due to the material differences of these classes,
that is reliably captured by the NIR channel.

\begin{figure}
\begin{center}
\begin{tabular}{cccc}
\hspace{-0.3cm}\includegraphics[height=.95cm]{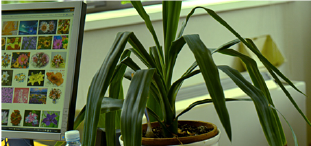}&
\hspace{-0.3cm}\includegraphics[height=.95cm]{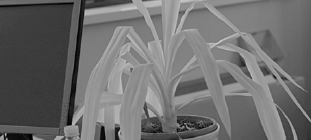}&
\hspace{-0.3cm}\includegraphics[height=.95cm]{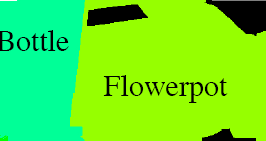}&
\hspace{-0.3cm}\includegraphics[height=.95cm]{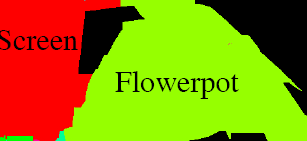}\\
\hspace{-0.3cm} \tiny RGB & \hspace{-0.3cm} \tiny NIR & \hspace{-0.6cm} \tiny 
$SIFT_{rgb}$ & \hspace{-0.5cm} \tiny  $SIFT_{rgbn}$
\end{tabular}
\caption{Due to very different material characteristics of vegetation,
  incorporating NIR information increases the accuracy of border detection.}
\label{fig:flowerpot}
\end{center}
\end{figure}

\textbf{Vegetation.} Incorporating NIR information in the segmentation task
significantly improves the result of conventional visible-only scenario in the
case of {\em flowerpot} because the very distinctive appearance of vegetation in
NIR images helps the classifier to correctly recognize the class and detect the
boundaries more accurately (see Figure~\ref{fig:flowerpot} for
illustration). The often higher contrast between this class and the background
makes it easy for the CRF model to more accurately detect the boundaries of this
class.

\subsection{Summary}

\begin{figure}[t]
\begin{center}
\begin{tabular}{cccc}
\hspace{-0.3cm}\includegraphics[height=2.7cm]{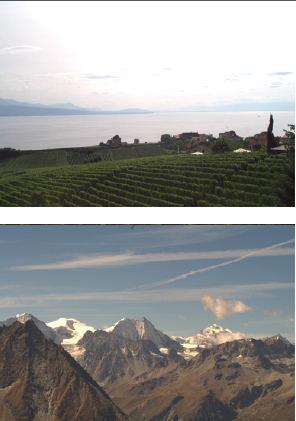}&
\hspace{-0.3cm}\includegraphics[height=2.7cm]{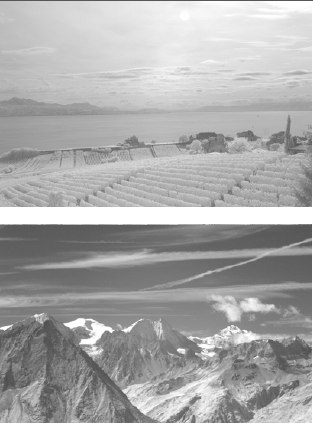}&
\hspace{-0.3cm}\includegraphics[height=2.7cm]{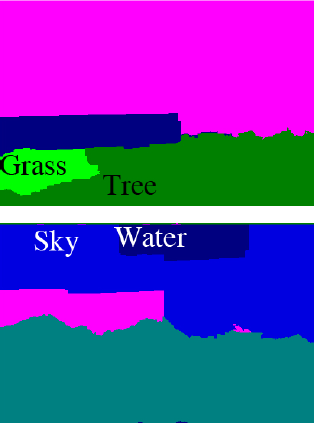}&
\hspace{-0.4cm}\includegraphics[height=2.7cm]{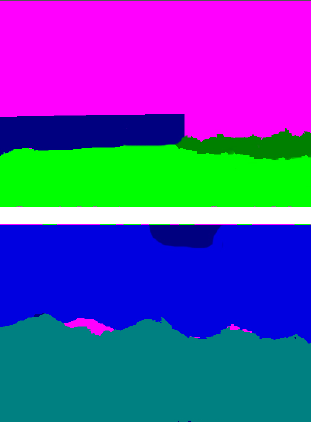}\\
\hspace{-0.3cm} \tiny RGB & \hspace{-0.3cm} \tiny NIR & \hspace{-0.7cm} \tiny 
$COL_{rgb}+SIFT_l$ & \hspace{-0.6cm} \tiny  $COL_{pc1234}+SIFT_n$
\end{tabular}
\caption{The material dependency characteristics of NIR images helps to distinguish more accurately between the classes of material with the same intrinsic color. Higher contrast in the NIR images in the sky makes $SIFT_n$ a more discriminative feature in distinguishing between $sky$ and $water$.}\label{fig:skygrass}
\end{center}
\end{figure}

Summarizing, we can say that classes such as {\em sky, grass, water} and {\em
  cloud} in the outdoor dataset are better recognized by their intrinsic
color. Capturing texture in a one-channel image and fusing it with the color
information improves the results, mostly by distinguishing between {\em grass}
and {\em tree}, or {\em sky} and {\em water}, where color is less
discriminative.  Figure~\ref{fig:skygrass} shows that incorporating
material-dependent NIR images in the decorrelated PCA space for the $COL$
descriptor and fusing it with $SIFT$ features on the NIR image help to recognize
such confusing classes.

By contrast, in the indoor scenes, most of the classes are man-made, and often
are of different colors such as cloths or handbags. Hence, the color is less
distinctive for recognizing the classes.  This can explain why $COL$ features
perform poorly compared to SIFT and similarly why multi-spectral SIFT
outperforms the late fusion of $COL$ and $SIFT$
features. Figure~\ref{fig:screencloth} shows examples where incorporating $COL$
gives poor results in the recognition of colorful classes. There, texture is
more intrinsic to the class, therefore multi-spectral-SIFT ($SIFT_{rgbn}$)
outperforms the late fusion of $COL$ and $SIFT$.


\section{Conclusion}
\label{sec:conclusion}
In this paper our aim was to explore the idea that NIR information, captured
from an ordinary digital camera, could be useful in semantic segmentation.

Therefore, we proposed to formulate the segmentation problem by using a CRF
model, and we studied ways to incorporate the NIR cue in the recognition part
and in the regularization part of our model. Considering the characteristics of
NIR images, we have defined color and SIFT features on different combinations of
the RGB and NIR channels. 

To evaluate this framework, we have introduced a novel
database of outdoor and indoor scene images, annotated at the pixel level, with
10 categories in the outdoor and 13 categories in the indoor scenes.

Through an extensive set of experiments, we have shown that integrating NIR as
additional information, along with conventional RGB images indeed improves the
segmentation results. We systematically studied the reasons for this improvement
by taking into consideration the material characteristics and properties of the
categories in the NIR wavelength range. In particular, the overall improvement
is due to a large improvement for certain classes whose response in the NIR
domain is particularly discriminant, such as \textit{water}, \textit{sky} or
\textit{screen}.

\vspace{0.5cm}
\noindent{\bf Acknowledgment.}
This work was supported by the Swiss National Science Foundation under grant
number 200021-124796/1 and by the Xerox foundation.

\bibliographystyle{plain}
\bibliography{template}

\end{document}